\documentclass{article} 
\usepackage{iclr2026_conference,times}

\usepackage{amsmath,amsfonts,bm}

\def\eqref#1{equation~\ref{#1}}

\def\1{\bm{1}}

\DeclareMathAlphabet{\mathsfit}{\encodingdefault}{\sfdefault}{m}{sl}
\SetMathAlphabet{\mathsfit}{bold}{\encodingdefault}{\sfdefault}{bx}{n}

\usepackage{hyperref}
\usepackage{url}
\usepackage{enumitem} 
\usepackage{tcolorbox}
\usepackage{siunitx}
\usepackage{booktabs}
\usepackage{multirow}
\usepackage[utf8]{inputenc} 
\usepackage[T1]{fontenc}    

\usepackage[table]{xcolor}
\usepackage{amsfonts}       
\usepackage{nicefrac}       
\usepackage{microtype}      

\usepackage{pifont}

\usepackage{makecell}
\usepackage{graphicx}
\usepackage{enumitem} 

\usepackage{siunitx}
\usepackage{amsmath}
\usepackage{wrapfig}
\usepackage{subcaption}
\usepackage[utf8]{inputenc}
\usepackage{CJKutf8}
\usepackage{algorithm}
\usepackage{algorithmic}
\usepackage{pifont}
\definecolor{BestInModule}{RGB}{235,245,255} 
\definecolor{BestOverall}{RGB}{235,255,235}
\definecolor{LightOrange}{RGB}{255, 248, 235}

\newtcolorbox[auto counter, number within=section]{namedbox}[2][]{
    colback=white,
    colframe=black,
    fonttitle=\bfseries,
    title=Box~\thetcbcounter: #2,
    #1
}

\title{SCI-Verifier:  
Scientific Verifier with Thinking}

\author{%
  Shenghe Zheng$^{1,2}$$^\dagger$, Chenyu Huang$^{3}$$^\dagger$, 
  Fangchen Yu$^{1,6}$, Junchi Yao$^{1,7}$, Jingqi Ye$^{1,8}$, \\ \textbf{Tao Chen}$^{3}$, \textbf{Yun Luo}$^{1}$, \textbf{Ning Ding}$^{1,5}$, \textbf{Lei Bai}$^{1}$,  \textbf{Ganqu Cui}$^{1}$, \textbf{Peng Ye}$^{1,4}$ \thanks{Corresponding Author~~~$^\dagger$Equal Contribution.~~~Project: \href{https://github.com/Zhengsh123/SCI-Verifier}{SCI-Verifier}}\\
  $^1$ Shanghai AI Laboratory \quad $^2$ Harbin Institute of Technology \quad $^3$ Fudan University \quad  \\ $^4$ CUHK \quad $^5$ Tsinghua University \quad $^6$ CUHK-Shenzhen \quad $^7$ UESTC \quad  $^8$ USTC
}

\iclrfinalcopy 

\begin{document}

\maketitle

\begin{abstract}
As large language models (LLMs) are increasingly applied to scientific reasoning, the complexity of answer formats and the diversity of equivalent expressions make answer verification a critical yet challenging task. Existing verification studies in scientific domains suffer from two major limitations: (a) the absence of systematic evaluation standards and insufficient disciplinary coverage, which hinders their comprehensive assessment; and (b) heavy reliance on cumbersome rule design or prompt engineering, which reduces their effectiveness in complex reasoning scenarios or limits their cross-disciplinary generalization.
To address these challenges, we propose solutions at both the data and model levels. On the data side, we construct \textbf{SCI-VerifyBench}, a cross-disciplinary benchmark covering mathematics, physics, biology, chemistry, and general scientific QA. The benchmark is built from real LLM responses and enhanced with domain-specific equivalence transformations that generate challenging and realistic data. Model-based and expert annotations ensure both quality and diversity, enabling rigorous evaluation of verification ability. On the model side, we emphasize the importance of reasoning for verification and introduce \textbf{SCI-Verifier}, a unified reasoning-augmented verifier for scientific domains. Through post-training, SCI-Verifier demonstrates strong logical reasoning and equivalence judgment capabilities while maintaining concise and stable outputs.
Together, SCI-VerifyBench and SCI-Verifier provide a principled framework for scientific verification, offering both systematic evaluation and practical pathways to enhance the reliability and applicability of LLMs in scientific domains.
\end{abstract}
\section{Introduction}\label{sec:intro}
As large language models (LLMs) become increasingly prevalent in scientific reasoning~\citep{yang2025qwen3, ren2025scientificintell, liu2024your, bai2025intern}, 
ensuring the reliability of their outputs has emerged as a critical challenge~\citep{chang2024survey,liu2025compassverifier}. Scientific reasoning often involves intricate multi-step processes and a wide range of equivalent answer formulations~\citep{chen2025reasoningerasurveylong}, posing substantial difficulties for answer verification. The essence of verification lies in accurately determining whether an output of LLM is equivalent to the reference answer, a task that serves both as the foundation for evaluating capabilities of LLMs and as a key bottleneck to further advancement~\citep{zhang2025reasoning, chen2025xverify, zhang2025compassjudger2}.

Despite recent progress, verification research in scientific domains still faces two major challenges. First, high-quality and systematic benchmarks are lacking. Existing benchmarks cover only a narrow range of scientific disciplines and fail to account for discipline-specific equivalence forms~\citep{liu2024rm, li2025verifybench, yan2025verifybench, chen2025xverify}, making it difficult to comprehensively evaluate verification capabilities. Second, current methods are limited in complex reasoning scenarios and lack cross-disciplinary applicability. Rule-based approaches rely on manually crafted templates and heuristics, which are labor-intensive and insufficient for complexity~\citep{math_verify_24_github, gao2021framework}. More recently, general LLMs or specialized verifiers leverage LLM generalization to achieve promising results~\citep{li2023making, liu2025compassverifier, zhang2025compassjudger2, luo2023chatgpt,liu2023evaluating}, but they require extensive prompt engineering, produce unstable outputs, and still struggle with complex reasoning and cross-disciplinary tasks. To overcome these limitations, we propose a solution from both the data and model perspectives, with the dual goals of establishing a systematic evaluation framework and designing a robust verifier tailored to scientific reasoning.
\begin{figure*}  
\centering  
\includegraphics[width=1 \textwidth]{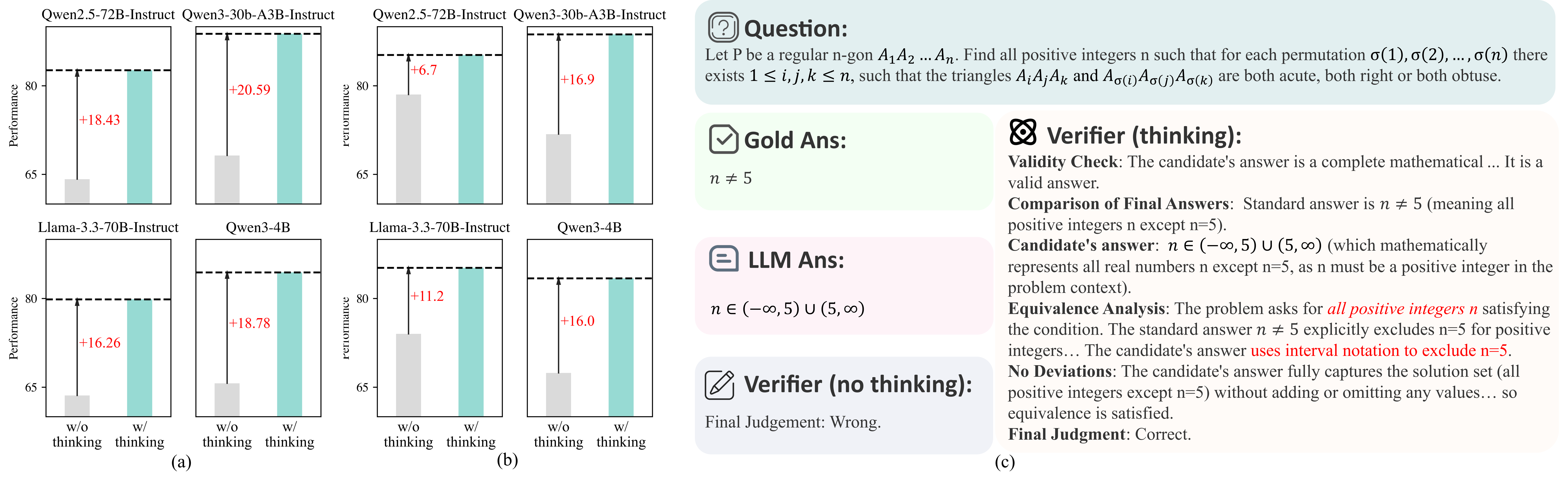} 
\vspace{-0.7cm}
\caption{(a) and (b) show the performance gains with reasoning enabled on VerifierBench and VerifyBench-hard, respectively; (c) presents a case where reasoning leads to the correct judgment.} 
\label{figure:motivation.}  
\vspace{-15pt}
\end{figure*} 

On the data side, we introduce SCI-VerifyBench, a cross-disciplinary and highly targeted scientific verification benchmark. We first collected over 100K responses from LLMs across mathematics, physics, biology, chemistry, and general scientific question-answering tasks, ensuring broad coverage of domains, question types, and answer formats. To increase both difficulty and specificity, we apply domain-specific equivalence transformation to some samples, enriching the test set with challenging cases. These transformations include, but are not limited to, formula rewrites, and logical equivalence substitutions, simulating error patterns that verifiers are likely to encounter in real-world scientific scenarios. Finally, by combining model-generated labels with expert human annotations, we ensured the quality and diversity of the data, making SCI-VerifyBench a benchmark that is both cross-disciplinary and rigorous for evaluating verification capabilities. In benchmark evaluations, we are surprised to find that reasoning abilities, which are often overlooked by many current specific verifiers, can significantly enhance the model's performance on scientific verification tasks.

At the model level, we propose SCI-Verifier, a reasoning-augmented verifier tailored for scientific domains.  In Fig.~\ref{figure:motivation.}, we highlight the critical role of reasoning in verification. Enabling Chain-of-Thought (CoT) across models consistently boosts judgment accuracy. Motivated by this, SCI-Verifier employs a two-stage post-training pipeline combining supervised fine-tuning and reinforcement learning to integrate logical reasoning into scientific verification, enabling it to handle complex equivalence judgments and multi-step reasoning frequently encountered in scientific tasks. It assesses equivalence from multiple perspectives while maintaining concise and stable outputs for practical deployment. Experimental results show that SCI-Verifier substantially improves accuracy on challenging and easily confusable samples compared to current verifiers and exhibits stronger cross-disciplinary generalization. Notably, the 8B version of SCI-Verifier achieves verification performance on par with the current state-of-the-art closed-source model GPT-5~\citep{Introducinggpt5}.

In summary, this work makes three key contributions as following:

\begin{enumerate}[labelsep = .5em, leftmargin = 0pt, itemindent = 1em]
    \vspace{-2pt}
    \item[$\bullet$] We propose a cross-disciplinary, high-challenge benchmark for scientific verification, SCI-VerifyBench, which covers mathematics, physics, biology, chemistry and general question-answer fields. Using real LLM responses and domain-specific equivalence transformations, it evaluates verification performance in complex scenarios and sets a unified standard for LLM assessment.
    \vspace{-2pt}
    \item[$\bullet$] We design a reasoning-enhanced high-performance scientific verifier, SCI-Verifier. By integrating logical reasoning via supervised post-training, SCI-Verifier gains the capability to perform complex equivalence judgments and conduct multi-step scientific reasoning, thereby significantly outperforming existing verification models across multiple domains.
    \vspace{-2pt}
    \item[$\bullet$]  Extensive experiments show that SCI-VerifyBench and SCI-Verifier together provide a precise evaluation framework and practical guidance for improving LLM capabilities, reliability, and reasoning in scientific domain, setting a new standard for cross-disciplinary verification research.
\end{enumerate}
\section{Related Works}
\textbf{Verification Benchmark.}
The unstructured LLM outputs makes the verification of the answers challenging, motivating efforts to construct benchmarks for evaluating the verifiers.
VAR~\citep{chen2025xverify} evaluates 19 LLMs on 24 datasets to train and assess xVerify. VerifyBench~\citep{li2025verifybench} covers general, logical, mathematical reasoning, and VerifyBench~\citep{yan2025verifybench} has 4,000 expert-level questions across STEM domains. VerifierBench~\citep{liu2025compassverifier} aggregates model outputs with manual meta-error analysis across math, science, knowledge, and general reasoning tasks.
Existing works face two main issues: (1)~pointless samples, such as multiple-choice questions that require no specially designed verifiers, and (2)~limited disciplinary coverage that restricts generalization assessment of scientific domain. To address these problems, we introduce SCI-VerifyBench, spanning mathematics, physics, chemistry, biology, and general QA, with filtered tasks and expert annotations to enable rigorous evaluation of scientific verification.

\textbf{Verification Models.} The verifier compensates for gaps in rule-based answer evaluation. xVerify~\citep{chen2025xverify} is efficient but lacks reasoning, limiting performance; General-Verifier~\citep{ma2025generalreasoner} has partial reasoning capabilities for cross-domain equivalence assessment. CompassVerifier~\citep{liu2025compassverifier} aims to provide efficient, high-performance, and robust answer verification using carefully designed error templates. Existing verifiers, constrained by limited reasoning capabilities, are inadequate for complex scientific reasoning, while using general models as verifiers requires careful prompt design with unstable outputs. Then, we propose SCI-Verifier, a reasoning-augmented scientific verifier offering strong reasoning with concise, stable outputs.

\textbf{Reward Models.} Reward models differ from verifiers in that they rank response quality, while verifiers assess correctness. Prior work primarily follows a discriminative paradigm, outputting a scalar score directly~\citep{ouyang2022training, snell2025scaling}. More recent approaches leverage reasoning capabilities to enhance reward model performance. For example, J1~\citep{whitehouse2025j1} proposes an RL framework for training Thinking-LLM-as-a-Judge models; Think-J~\citep{huang2025thinkj} introduces offline and online RL-based methods for judgment thinking optimization; Compass-Judger2~\citep{zhang2025compassjudger2} uses verifiable rewards and rejection sampling to guide critical reasoning, improving robustness and generalization. Despite these advances, reward models aim to rank response quality rather than verify correctness, and this difference in objective introduces new challenges for data construction and training strategies.

\section{SCI-VerifyBench}\label{sec:varitasbench}
\begin{wraptable}{r}{0.52\textwidth} %
\vspace{-15pt}
\centering
\caption{Comparison of verification benchmarks.}\label{tab:bench:compare}
\vspace{-8pt}
\scalebox{0.7}{
\begin{tabular}{lcccc}
    \hline
        &\multirow{2}{*}{Scale} & \multirow{2}{*}{Domains} & \multirowcell{2}{{Equivalence}\\ {Transformation}} & \multirowcell{2}{{Difficulty}\\ {Control}}\\
        & & & & \\ \midrule
        VerifyBench & 2000 & 3 & \ding{55} & \ding{55} \\ 
        VerifyBench-hard & 1000 & 3 & \ding{55} & \ding{51} \\ 
        VerifierBench & 2817 & 4 & \ding{55} & \ding{55} \\ \midrule
        SCI-VerifyBench & 2500 & \textbf{5} & \ding{51}& \ding{51} \\ \bottomrule
    \end{tabular}
    }
    \vspace{-5pt}
\end{wraptable}
Current research on scientific verification faces a major bottleneck in the lack of comprehensive and rigorous benchmarks, which creates blind spots in evaluating LLMs’ scientific reasoning capabilities and guiding their training. To address this problem, we construct SCI-VerifyBench, a systematic cross-disciplinary benchmark covering mathematics, physics, chemistry, biology, and general scientific QA, designed to comprehensively evaluate the verification abilities of verifiers. We first present the characteristics of SCI-VerifyBench in Sec.~\ref{sec:varitasbench:analysis}, followed by the construction pipeline in Sec.~\ref{sec:varitasbench:collection}.

\begin{figure*}  
\centering  
\includegraphics[width=1 \textwidth]{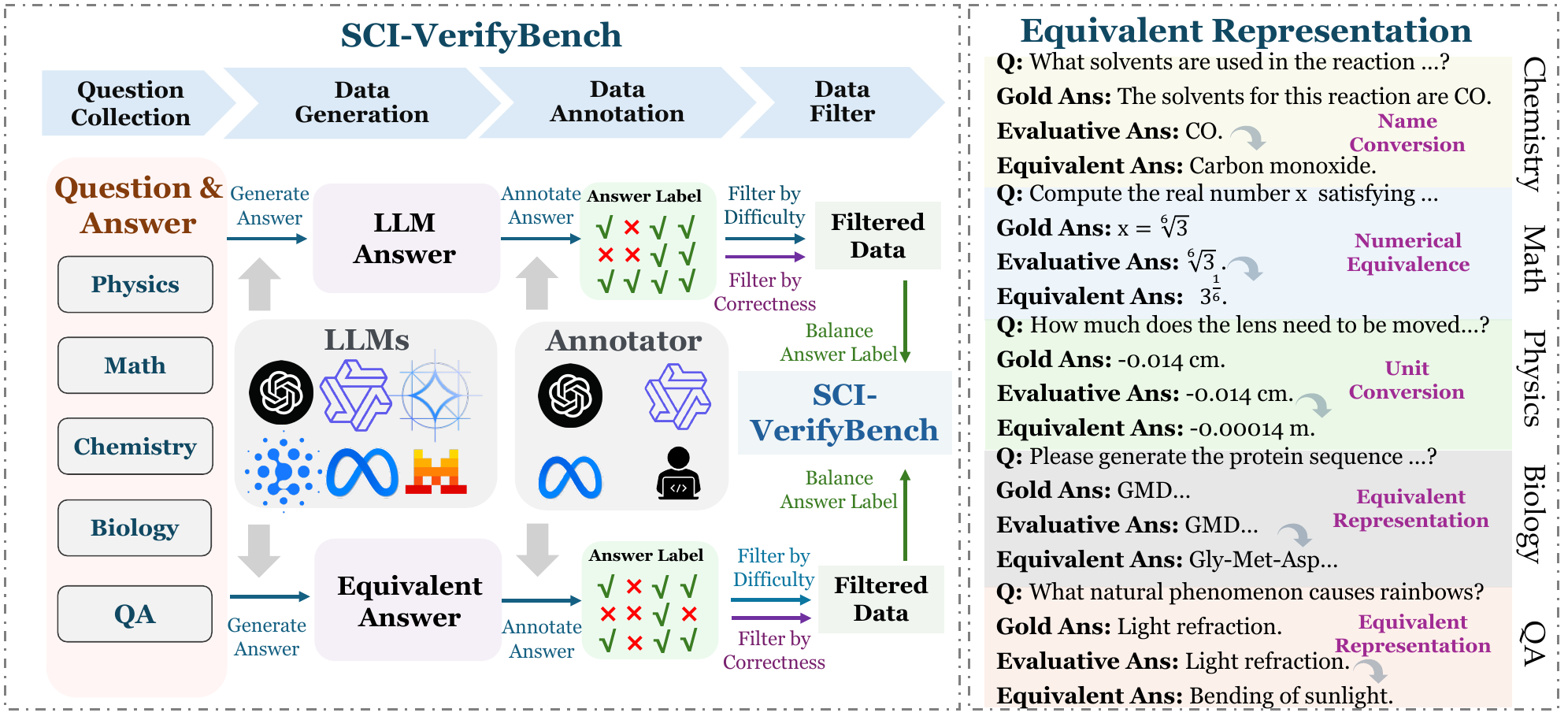} 
\vspace{-0.3cm}
\caption{SCI-VerifyBench construction pipeline (left) and cases of Equivalent Representation (right).} 
\label{figure:pipeline.}  
\vspace{-10pt}
\end{figure*} 
\begin{wraptable}{r}{0.4\textwidth} %
\vspace{-20pt}
\centering
\caption{Benchmark Statistic.}\label{tab:bench:statics}
\vspace{-8pt}
\scalebox{0.8}{
\begin{tabular}{ll}
    \toprule
        Statistic & Number \\ \midrule
        Total Data (each domain) & 500 \\ 
        Real QA (each domain) & 350 \\ 
        Synthetic QA (each domain) & 150 \\ \midrule
        Average Answer Tokens & 24.98 \\ 
        Average Response Tokens & 2980.43 \\ \bottomrule
    \end{tabular}
    }
    \vspace{-10pt}
\end{wraptable}
\subsection{Data Overview}\label{sec:varitasbench:analysis}

We construct SCI-VerifyBench to assess verifiers’ scientific verification capabilities. In this part, we compare it with existing benchmarks and present static data analyses. Tab.~\ref{tab:bench:compare} highlights the key differences, showing that SCI-VerifyBench spans a wider range of domains, incorporates more challenging yet commonly encountered equivalence transformations, and applies difficulty control mechanisms to better reflect realistic scientific reasoning while increasing overall task complexity. Tab.~\ref{tab:bench:statics} presents the static analysis results, with further details provided in Appendix~\ref{sec:appendix:statics}.

\subsection{Data Collection}\label{sec:varitasbench:collection}

\textbf{Question–Answer Data Generation.} Comprehensive scientific verification requires coverage of diverse question areas, answer types, and response formats. To achieve this, we collect over 15k question-answer pairs across mathematics~\citep{omini-math}, physics~\citep{PHYSICS}, chemistry~\citep{chembench}, biology~\citep{bioinformatics}, and general QA~\citep{mmlupro,simpleqa}, and generate 100K+ responses using eight models of varying scales, while controlling response length. This design allows the verifier to adapt to different question and answer styles. The prompts used for data generation are provided in the Appendix~\ref{sec:appendix:statics:generation}.

\textbf{Synthetic Data Generation.} Current verifiers perform well on responses identical to reference answers. However, even carefully trained specialized verifiers can fail when confronted with complex, domain-specific equivalence transformations, which are common in real-world scientific reasoning. Fig.~\ref{figure:pipeline.} illustrates typical equivalence transformations across mathematics, physics, chemistry, biology, and general QA. To address this, we select 500 representative questions from each domain that allow for answer equivalence transformations and generate five equivalent answers for each using the methodology described in Appendix~\ref{sec:appendix:statics:generation}. During this process, five LLMs assist in assessing the quality of generated equivalences. If the equivalence is clearly invalid and multiple models agree, the sample is discarded and regenerated. This approach increases the challenge for verifiers and closely simulates the diverse answer forms encountered in realistic scientific reasoning scenarios.

\textbf{Data Annotation.} The previous two pathways produced both real and synthetic question-answer data. To ensure annotation quality, we adopt a hybrid approach combining LLMs and human experts. First, five LLMs evaluate the accuracy of generated or synthetic answers against the reference answers as shown in Appendix~\ref{sec:appendix:statics:eval}. To reduce human effort and maintain dataset difficulty, we retain only samples where the five LLMs disagree. From the data in the two methods described above, we select 2,500 samples with the highest model disagreement (500 per domain) for human annotation. Each sample is assessed by at least two experts with a bachelor’s degree or higher, and answers are considered equivalent if they can be transformed into each other. In cases of disagreement, a third expert is consulted, ensuring labels reflect true equivalence while maintaining diversity.

\textbf{Data Filter.} Using the above procedure, we obtained a dataset comprising 5,000 human-annotated samples and a large collection of LLM-annotated samples. We partitioned the data into a training set and a test set, with the latter corresponding to SCI-VerifyBench. For the test set, we sampled 350 real LLM responses and 150 equivalence-based synthetic responses per domain. The selection criterion required full agreement among human experts, while samples with disagreement between human experts and LLMs are preferentially included to increase difficulty. This process results in a test set of 2,500 samples in total. The remaining non-overlapping data formed the training set, where only a portion was human-annotated and the majority relied on LLM annotations. Samples with substantial disagreement among LLMs were filtered out to ensure label reliability, yielding a training set of 14K samples. Both the training and test sets can be expressed in the following format:
\begin{equation}
    D=\{(q_i,a_i,r_i,l_i)\}_{i=1}^N,
\end{equation}
where $q_i$ denotes the question, $a_i$ denotes the reference answer, $r_i$ denotes the response whose correctness needs to be evaluated, and $l_i$ denotes the label, which can be either \texttt{true} or \texttt{false}.

\section{SCI-Verifier}\label{sec:r_verifier}

In this section, we develop a reasoning-augmented verifier for scientific verification.  First, Sec.~\ref{sec:r_verifier:motivation} presents the motivation and necessity for incorporating reasoning capabilities into scientific verification. Then, Sec.~\ref{sec:r_verifier:post} describes the approaches for integrating reasoning through supervised fine-tuning (SFT) and reinforcement learning (RL). This design emulates human step-by-step reasoning and improves verification reliability and robustness.

\subsection{Motivation}\label{sec:r_verifier:motivation}
We begin by motivating the introduction of reasoning capabilities into scientific verification. While Chain-of-Thought (CoT) has been widely recognized for enhancing model performance across various domains, most existing verifier studies have overlooked this aspect. As shown in Fig.~\ref{figure:motivation.}, we evaluate models of different scales under two conditions: outputting only the final answer versus producing intermediate reasoning before the answer. The results demonstrate that reasoning brings substantial gains in scientific verification, largely because responses of scientific questions are inherently complex and often involve multiple equivalent forms. Reasoning is thus essential for assessing equivalence from different perspectives. Building on this insight, we argue that optimizing reasoning for scientific verification can significantly boost verifier performance. Next, we introduce SCI-Verifier, a unified verifier designed to deliver concise yet powerful reasoning capabilities.

\subsection{Post-Training}\label{sec:r_verifier:post}
In this section, we present our approach to enhancing the reasoning ability of models for scientific verification. While reasoning is essential, the nature of verification requires reasoning paths to be as concise as possible to minimize resource consumption. Accordingly, we aim for a lightweight verifier with short and stable outputs. Based on these characteristics, we utilize a two-stage post-training paradigm that combines supervised fine-tuning (SFT) with reinforcement learning (RL).

\textbf{Supervised Fine-Tuning (SFT).} In this stage, we employ large models with rejection sampling to generate a diverse set of reasoning paths in a structured format as shown in Appendix~\ref{sec:appendix:statics:eval}. We then perform strict filtering to retain only valuable and concise traces. For reasoning models, we keep only the conclusive summary, while for non-reasoning models, we discard overly long or unstructured responses. The filtered reasoning paths are used to fine-tune a smaller model, effectively injecting the essential reasoning ability with minimal overhead. The training objective is defined as follows:
\begin{equation}
\mathcal{L}_{\mathrm{SFT}}(\theta) = - \mathbb{E}_{(x, y) \sim \mathcal{D}_{\text{SFT}}} 
\Big[ \log \pi_{\theta}(y \mid x) \Big],
\end{equation}
where $\mathcal{D}_{\text{SFT}}$ denotes the curated training dataset of high-quality reasoning traces.
Unlike SFT in other domains like mathematics or physics where the focus is mostly on output formatting, verification SFT centers on transferring domain-specific verification knowledge to small models, which is essential for developing concise and useful reasoning for verification.

\textbf{Reinforcement Learning (RL).} After SFT, the model acquires basic verification capability and the ability to generate outputs in a fixed format. However, it inevitably faces the challenges of overfitting. To improve generalization, we adopt DAPO~\citep{yu2025dapo}, which is a refined GRPO~\citep{deepseek-math} method.
Within this framework, training samples are dynamically filtered. Overly simple samples fail to provide meaningful learning signals, while overly difficult samples may destabilize training. To further encourage concise reasoning, we incorporate a length penalty into the reward function. The overall objective function for training is to maximize the following expression: 
\begin{align}
\mathcal{J}_{\text{DAPO}}(\theta) 
&= \mathbb{E}_{(q, a) \sim \mathcal{D}, \; \{o_i\}_{i=1}^G \sim \pi_{\theta_{\text{old}}}(\cdot|q)} \Bigg[ 
     \frac{1}{|\{o_i\}|} \sum_{i=1}^{G} \sum_{t=1}^{|o_i|} \nonumber \\
&\quad \min \Big( r_{i,t}(\theta)\, \hat{A}_{i,t}, \;
     \text{clip}\big(r_{i,t}(\theta), \; 1 - \epsilon_{\text{low}}, \; 1 + \epsilon_{\text{high}}\big)\, \hat{A}_{i,t} \Big) 
\Bigg] 
\label{eq:dapo_overall}
\end{align}

The advantage function $\hat{A}_{i,t}$ is calculated from the final reward $R_i$: \begin{equation} \hat{A}_{i,t} = \frac{R_i - \text{mean}(\{R_j\}_{j=1}^G)}{\text{std}(\{R_j\}_{j=1}^G)} \label{eq:advantage} 
\end{equation}  
The final reward $R_i$ is a sum of the alignment reward $R_{\text{align},i}$ and a overlong penalty $P_{\text{overlong},i}$: 
\begin{equation}     R_i = R_{\text{align},i} + P_{\text{overlong},i}     \label{eq:final_reward} \end{equation} 
where the overlong penalty is defined as: \begin{equation} 
P_{\text{overlong},i} =  \begin{cases}     0 & \text{if } |o_i| \leq L_{\text{max}} \\     -\frac{|o_i| - L_{\text{max}}}{L_{\text{buffer}}} \cdot \lambda_{\text{penalty}} & \text{if } L_{\text{max}} < |o_i| \leq L_{\text{max}} + L_{\text{buffer}} \\     -\infty & \text{if } |o_i| > L_{\text{max}} + L_{\text{buffer}} \end{cases} \label{eq:overlong_penalty} 
\end{equation}
Here, $|o_i|$ is the length of the response, $L_{\text{max}}$ is the maximum allowed length, $L_{\text{buffer}}$ is the overlong buffer length, and $\lambda_{\text{penalty}}$ is the penalty weight.
Since verification is a binary classification task, imbalanced data may lead the model to rely on label priors instead of reasoning. To address this, we rebalance the dataset during RL training to ensure equal positive and negative examples.

Through this two-stage post-training paradigm, we obtain a verification model with concise reasoning ability, applicable across domains for scientific answer validation. This approach not only improves the reliability of model capability evaluation, but also provides a reward function with clear semantics, thereby facilitating the training of stronger reasoning-oriented language models.
\section{Experiments}\label{sec:exp}

\begin{table}[!t]
    \centering
    \caption{
   Performance of different verifiers on SCI-VerifyBench. Specialized verifiers use default prompts, while all other models are allowed to use reasoning.
    }
    \vspace{-5pt}
    \label{tab:overall_eval_results}
    \resizebox{\columnwidth}{!}{
    \begin{tabular}{@{}l|ccccc|c|r@{}}
        \toprule
        \textbf{Models} & \textbf{Math}&\textbf{Physics}&\textbf{Chemistry}&\textbf{Biology}&\textbf{QA}&\textbf{Total}&\textbf{Avg. Token}\\
        
\rowcolor{LightOrange}
\multicolumn{8}{c}{Closed-source Models}\\

GPT-5~\citep{Introducinggpt5} &
82.60&
74.60&
\cellcolor{BestOverall}\textbf{88.00}&
89.00&
\cellcolor{BestOverall}\textbf{90.40}&
84.92&
384.59\\

Gemini-2.5-Flash~\citep{comanici2025gemini}
&
80.40&
68.80&
84.00&
88.80&
89.00&
82.20&
478.48\\

o4-mini~\citep{IntroducingOpenAIO3}
&
78.40&
69.80&
83.20&
88.00&
89.20&
81.72&
437.27\\

\rowcolor{LightOrange}
\multicolumn{8}{c}{Open-source Instruct models}\\

Qwen2.5-72B-Instruct~\citep{qwenQwen25TechnicalReport2025}&
75.60&
62.40&
76.80&
80.80&
79.40&
75.00&
400.40\\
Qwen3-30B-A3B-Instruct-2507~\citep{qwenQwen25TechnicalReport2025}&
75.40&
63.60&
79.80&
89.80&
82.20&
78.16&
684.58\\

LLaMa-3.3-70B-Instruct~\citep{grattafioriLlama3Herd2024}&
76.60&
67.40&
78.80&
84.60&
85.80&
78.64&
364.36\\
\rowcolor{LightOrange}
\multicolumn{8}{c}{Open-source Reasoning Models}\\

Qwen3-4B~\citep{yang2025qwen3}&
73.20&
59.80&
79.40&
81.00&
78.80&
74.44&
1466.92\\
Qwen3-8b~\citep{yang2025qwen3}&
75.00&
61.60&
79.20&
81.80&
75.00&
74.52&
1033.07\\
GPT-oss-20b~\citep{openai2025gptoss120bgptoss20bmodel}&
71.00&
54.20&
80.40&
80.40&
65.60&
70.32&
522.55\\
Qwen3-30B-A3B-Thinking-2507~\citep{yang2025qwen3}&
76.20&
64.40&
83.00&
82.40&
78.80&
76.96&
1714.66\\

GPT-oss-120B~\citep{openai2025gptoss120bgptoss20bmodel}&
78.80&
67.80&
\cellcolor{BestInModule}\underline{84.20}&
86.00&
89.00&
81.16&
110.21\\

Qwen3-235B-A22B~\citep{yang2025qwen3}&
76.40&
61.00&
81.80&
85.20&
73.80&
75.64&
4601.16\\
\rowcolor{LightOrange}
\multicolumn{8}{c}{Specific Verifiers}\\

xVerify-8B~\citep{chen2025xverify}&
72.80&
55.60&
80.80&
83.60&
83.00&
75.16&
1.00
\\

CompassVerifier-3B~\citep{liu2025compassverifier} &
75.00&
66.40&
82.00&
84.20&
77.00&
76.92&
192.00\\
CompassVerifier-7B~\citep{liu2025compassverifier} & 
76.00&
65.00&
80.40&
81.00&
79.00&
76.28&
162.34\\
CompassVerifier-32B~\citep{liu2025compassverifier} &
77.40&
67.60&
81.60&
80.60&
82.80&
78.00&
212.04
\\
\rowcolor{LightOrange}
\multicolumn{8}{c}{Ours}
\\

SCI-Verifier-4B&
\cellcolor{BestInModule}\underline{86.20}&
\cellcolor{BestInModule}\underline{77.40}&
83.20&
\cellcolor{BestInModule}\underline{90.80}&
\cellcolor{BestInModule}\underline{89.40}&
\cellcolor{BestInModule}\underline{85.40}&
485.13
\\

SCI-Verifier-8B&
\cellcolor{BestOverall}\textbf{87.60}&
\cellcolor{BestOverall}\textbf{79.60}&
80.60&
\cellcolor{BestOverall}\textbf{94.40}&
89.20&
\cellcolor{BestOverall}\textbf{86.28}&
490.66
\\

        \bottomrule
    \end{tabular}
    }
    \vspace{-15pt}
\end{table}

\begin{table}[!t]
    \centering
    \caption{
   Performance of different verifiers on VerifierBench and VerifyBench-Hard. Specialized verifiers use default prompts, while all other models are allowed to use reasoning.
    }
    \vspace{-5pt}
    \label{tab:other_eval_result}
    \resizebox{\columnwidth}{!}{
    \begin{tabular}{@{}l|ccr|ccr@{}}
        \toprule
        \textbf{Models} & \multicolumn{3}{c}{\textbf{VerifierBench}}&\multicolumn{3}{c}{\textbf{VerifyBench-Hard}}\\
        &Acc.&F1&Avg. Token&Acc.&F1&Avg. Token\\
\rowcolor{LightOrange}
\multicolumn{7}{c}{Closed-source Models}\\

GPT-5~\citep{Introducinggpt5}&
91.80&
90.48&
203.45&
90.40&
85.34&
245.64\\
Gemini-2.5-Flash~\citep{comanici2025gemini}&
87.63&
87.56&
265.49&
87.70&
83.65&
302.65
\\


\rowcolor{LightOrange}
\multicolumn{7}{c}{Open-source Instruct models}\\

Qwen2.5-72B-Instruct~\citep{qwenQwen25TechnicalReport2025}&
82.61&
81.67&
550.73&
85.20&
81.31&
381.27\\
Qwen3-30B-A3B-Instruct-2507~\citep{qwenQwen25TechnicalReport2025}&
88.78&
88.88&
972.30&
88.70&
85.03&
810.24\\

LLaMa-3.3-70B-Instruct~\citep{grattafioriLlama3Herd2024}&
79.84&
79.00&
398.99&
85.20&
81.10&
382.04\\
\rowcolor{LightOrange}
\multicolumn{7}{c}{Open-source Reasoning Models}\\

Qwen3-4B~\citep{yang2025qwen3}&
84.42&
84.54&
2119.87&
83.40&
78.80&
1755.61\\
Qwen3-8b~\citep{yang2025qwen3}&
85.55&
85.56&
1857.95&
84.40&
79.58&
1588.45\\
GPT-oss-20B~\citep{openai2025gptoss120bgptoss20bmodel}&
83.36&
83.73&
523.10&
85.90&
80.36&
328.50\\
Qwen3-30B-A3B-Thinking-2507~\citep{yang2025qwen3}&
90.42&
90.05&
2438.52&
88.60&
84.92&
2226.46\\


Qwen3-235B-A22B~\citep{yang2025qwen3}&
88.36&
88.01&
5044.43&
86.80&
82.26&
4690.13\\
\rowcolor{LightOrange}
\multicolumn{7}{c}{Specific Verifiers}\\

xVerify-8B~\citep{chen2025xverify}&
78.03&
75.53&
1.00&
83.20&
79.60&
1.00
\\

CompassVerifier-3B~\citep{liu2025compassverifier} &
82.39&
83.37&
1.00&
86.60&
84.16&
1.00\\
CompassVerifier-7B~\citep{liu2025compassverifier} & 
85.56&
84.83&
1.00&
87.50&
84.13&
1.00\\
CompassVerifier-32B~\citep{liu2025compassverifier} &
89.88&
88.91&
1.00&
88.30&
85.86&
1.00\\
\rowcolor{LightOrange}
\multicolumn{7}{c}{Ours}
\\

SCI-Verifier-4B&
\cellcolor{BestInModule}\underline{92.37}&
\cellcolor{BestInModule}\underline{92.01}&
703.47&
\cellcolor{BestInModule}\underline{88.90}&
\cellcolor{BestInModule}\underline{85.98}&
470.26
\\

SCI-Verifier-8B&
\cellcolor{BestOverall}\textbf{93.01}&
\cellcolor{BestOverall}\textbf{93.06}&
636.53&
\cellcolor{BestOverall}\textbf{90.30}&
\cellcolor{BestOverall}\textbf{87.45}&
393.61
\\

        \bottomrule
    \end{tabular}
    \vspace{-15pt}
    }
\end{table}
\subsection{Baselines and Setup}\label{sec:exp:setup}
We conduct a systematic evaluation of SCI-VerifyBench on SCI-Verifier-4B and 8B, which are trained from Qwen3-4B-Base~\citep{yang2025qwen3} and Qwen3-8B-Base~\citep{yang2025qwen3}, respectively. In addition, we benchmark on two established datasets: VerifierBench~\citep{liu2025compassverifier} and VerifyBench-hard~\citep{yan2025verifybench}. The baselines cover four categories: (1) closed-source models, (2) open-source instruct models, (3) open-source reasoning models, and (4) specialized verifiers. Details are provided in Appendix~\ref{sec:appendix:statics:eval}. For evaluation, we report Accuracy on SCI-VerifyBench, since positive and negative samples are balanced by construction. On VerifierBench and VerifyBench-hard, we additionally report F1 score alongside Accuracy. In all cases, higher values indicate stronger verification performance.
\subsection{Evaluation And Analysis of SCI-VerifyBench}\label{sec:exp:varitas}
In this part, we present and analyze the evaluation results on SCI-VerifyBench. Tab.~\ref{tab:overall_eval_results} reports the performance of both closed-source and open-source models on SCI-VerifyBench. This comprehensive evaluation enables us to compare the verification capabilities of LLMs of different types and scales under the same settings. We then provide a detailed analysis of the experimental results.

\textbf{Open-source models are gradually closing the gap with proprietary models, yet a noticeable performance gap remains. }On the verification task, many open-source models have approached the performance of closed-source models, including specialized verifiers, but proprietary models still maintain an edge. For instance, GPT-5 outperforms current open-source models by more than 5\%. Notably, our proposed SCI-Verifier achieves performance comparable to GPT-5 on the scientific verification task, which confirms the effectiveness of the proposed verifier.

\textbf{Reasoning models and chat models do not exhibit significant differences on this task.} On this task, reasoning models show no clear advantage over chat models. We attribute this to the fact that, unlike challenging problems such as IMO-level mathematics, scientific verification tasks are straightforward, requiring domain-specific knowledge and only brief reasoning. Since both model types share similar priors and lack reasoning-specific optimization, performance gains are limited. This observation underscores the need for reasoning tailored to the unique characteristics of verification tasks.

\begin{wrapfigure}[16]{r}{0.45\textwidth} 
  \vspace{-1.8em}  
  \centering
  \includegraphics[width=0.9\linewidth]{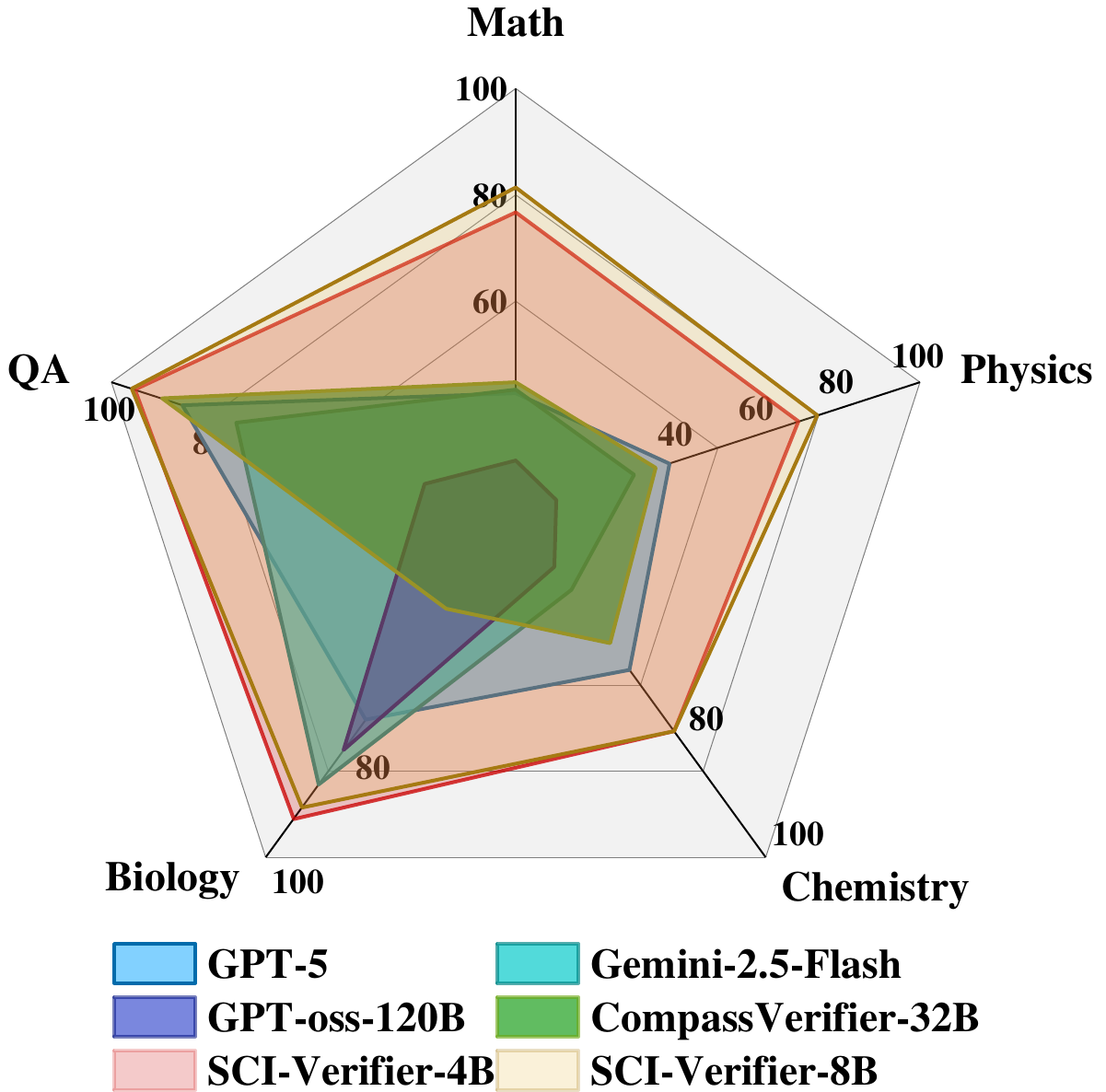}
  \vspace{-3pt}
  \caption{Evaluation on Equivalent Answer.} 
  \label{fig:equiv_results}
\end{wrapfigure}

\textbf{Equivalence-based answers poses significant challenges for current LLMs.} As shown in Fig.~\ref{fig:equiv_results}, on our equivalence-augmented test set derived from SCI-VerifyBench, even state-of-the-art GPT-5 models perform poorly, with scores dropping below 50\% in mathematics and physics. This highlights a clear deficiency in handling complex equivalence transformations. Remarkably, our SCI-Verifier, in both its 4B and 8B configurations, achieves substantially higher performance on the same tests, owing to targeted optimization for this challenge. These results provide strong evidence for the effectiveness of integrating reasoning capabilities specifically tailored for equivalence verification.

\begin{figure*}  
\centering  
\includegraphics[width=1 \textwidth]{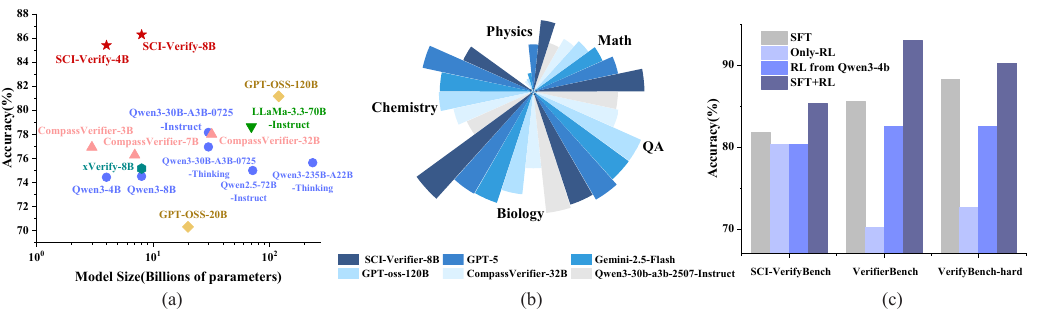} 
\vspace{-0.7cm}
\caption{(a) Performance on SCI-VerifyBench versus model size.
(b) Difficulty comparison across domains in SCI-VerifyBench.
(c) Ablation study of training methods.} 
\label{figure:combine_1.}  
\vspace{-10pt}
\end{figure*}

\textbf{Model scale does not have a decisive impact on results.} Experiments across model scales show that scaling up the model does not consistently improve performance as shown in Fig.~\ref{figure:combine_1.}(a). We hypothesize this is because the verification primarily depends on prior knowledge to assess answer equivalence. Since current models are not optimized for this task, improvements in model capacity do not translate into enhanced verification performance.

\textbf{Task characteristics across domains lead to domain-dependent performance differences.} As shown in Tab.~\ref{tab:overall_eval_results} and Fig.~\ref{figure:combine_1.}(b), performance varies across disciplines and exhibits consistent trends across different models. Scores in mathematics and physics are lower than in other subjects, mainly due to the complex transformations required in these domains, such as factorization, and Taylor expansions, which introduce greater task subtlety. In contrast, judgments in other disciplines are more straightforward once prerequisite knowledge is available. These results highlight the need for verifiers tailored to each discipline's characteristics.

\subsection{Evaluation And Analysis of SCI-Verifier}\label{sec:exp:verifier}
\textbf{Generalization of SCI-Verifier.} We conduct experiments on our SCI-VerifyBench and two existing verification benchmarks, VerifierBench and VerifyBench-Hard as shown in Tab.~\ref{tab:other_eval_result}. The results demonstrate that, both sizes of SCI-Verifier achieve strong performance even at small sizes, reaching levels comparable to the state-of-the-art closed-source model GPT-5. Meanwhile, Fig.~\ref{fig:equiv_results} demonstrates the strong capability of SCI-Verifier in judging equivalence transformations. The consistent advantage of SCI-Verifier across all three benchmarks indicates its strong verification ability and generalization capability across tasks. Notably, on SCI-VerifyBench, SCI-Verifier outperforms current open-source models in all disciplines, further validating its cross-disciplinary generalization in verification.

\begin{wraptable}{r}{0.65\textwidth} %
\vspace{-10pt}
\centering
\caption{Comparison of model robustness across different prompts. \textit{our}: default prompt; \textit{other}: modified prompt.}\label{tab:bench:robust}
\vspace{-8pt}
\scalebox{0.8}{
\begin{tabular}{l|cc|cc}
    \toprule
        \textbf{Models}&\multicolumn{2}{c|}{\textbf{SCI-VerifyBench}}&\multicolumn{2}{c}{\textbf{VerifyBench-Hard}} \\ 
        & our &other &our &other\\ \midrule
        Qwen3-30B-A3B-Instruct-2507&
        78.16&
        76.72&
        88.70&
        75.40\\ 
        GPT-oss-20b&
        70.32&
        78.08&
        85.90 &
        79.50\\ 
        Qwen3-235B-A22B&
        75.64&
        77.00&
        86.80&
        81.00\\ 
        CompassVerifier-3B&
        76.92&
        74.52&
        86.60&
        79.30\\ 
        CompassVerifier-7B&
        76.28&
        76.44&
        87.50&
        80.30\\ 
        CompassVerifier-32B&
        78.00 &
        81.00&
        88.30&
        84.70\\ \midrule
        SCI-Verifier-4B&
        85.40&
        84.90&
        88.90&
        88.30\\ 
        SCI-Verifier-8B&
        86.28&
        85.50&
        90.30&
        89.70\\ \bottomrule
    \end{tabular}
    }
    \vspace{-7pt}
\end{wraptable}
\textbf{Prompt Robustness of SCI-Verifier.} We investigate the robustness of SCI-Verifier to prompt variations, a property that is crucial for real-world applications where prompts must often be adapted to user requirements~\citep{liu2023pre}. We evaluate multiple models on three benchmarks using both our proposed CoT prompt and the xVerify prompt (modified to allow reasoning for alignment purposes). Details are provided in Appendix~\ref{sec:appendix:statics:eval}, and the results are summarized in Tab.~\ref{tab:bench:robust}. From these results, we draw two key conclusions: (1) SCI-Verifier exhibits strong robustness to prompt modifications, maintaining competitive performance even when the prompt differs from those seen during training; and (2) general models are considerably more sensitive to prompt variations in verification tasks, largely because they lack an intrinsic notion of answer equivalence and must instead rely on contextual cues. Notably, this sensitivity tends to diminish as model size increases.
\subsection{Ablation Study}\label{sec:exp:ablation}

\begin{figure*}  
\centering  
\includegraphics[width=1 \textwidth]{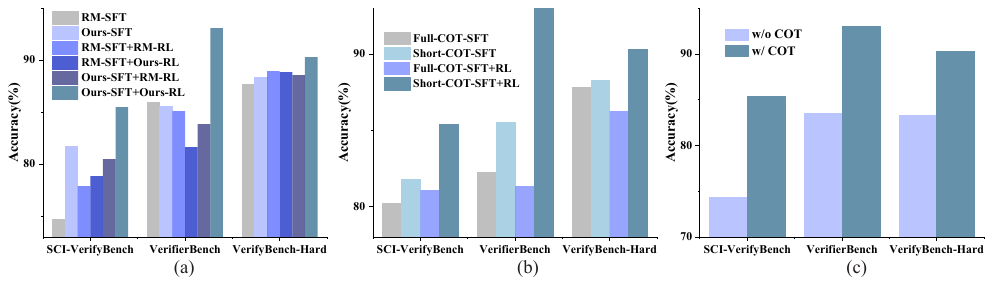} 
\vspace{-0.7cm}
\caption{(a) Ablation study of training data impact.
(b) Ablation study of SFT distillation methods.
(c) Ablation study of training with CoT in scientific verification.} 
\label{figure:combine_2.}  
\vspace{-10pt}
\end{figure*}

\textbf{Training Methods.} In this section, we analyze the contribution of each component in our two-stage training framework. We conduct experiments on both 4B and 8B models, with results shown in Fig.~\ref{figure:combine_2.}(a). We observe that applying SFT on the base model alone already yields relatively strong performance on verification tasks. Starting RL from a reasoning model also achieves competitive results, whereas directly applying RL on the Base model performs poorly. This may be due to the absence of SFT warm-up, where the Base model requires a large amount of training data to acquire targeted reasoning abilities. By contrast, combining SFT with RL leads to consistently superior performance, particularly in terms of generalization across different datasets. These findings highlight that both stages of the proposed training framework are indispensable.

\textbf{Training Data.}   In this part, we evaluate the quality of our constructed training dataset by comparing it with a commonly used dataset~(RM)~\citep{zhao2025one} in the Reward Model domain. Using Qwen3-4B-Base as the initial model, we conduct experiments with both datasets under SFT and SFT+RL settings, and the detailed results are presented in Fig.~\ref{figure:combine_2.}(b). The results show that our dataset consistently enables the model to achieve strong performance across three benchmarks, whether used for SFT or RL. This demonstrates the high quality of our data, from which the model can learn richer distributional information about the verification task. The RM dataset also yields reasonable performance under SFT, mainly because of its large scale with more than 180K samples. However, its effectiveness under RL is limited since the heterogeneous quality within such a large dataset slows down model improvement, which makes data filtering necessary in practice. These findings confirm that our constructed training dataset, like our test data, is of high quality and reliability.

\textbf{Distillation Data.}   We investigate the effectiveness of our proposed short CoT distillation. Specifically, we compare the outcomes of distilling complete CoT versus short CoT, with results presented in Fig.~\ref{figure:combine_2.}(b). The findings reveal that distilling complete CoT not only fails to improve performance but also substantially increases output length, rendering it impractical. We attribute this to the nature of the verification task, which is relatively simple and does not require long reasoning chains. Instead, concise reasoning from fixed perspectives is sufficient to achieve strong performance. Therefore, distilling short reasoning traces during the SFT stage is both a reasonable and efficient choice.

\textbf{Inference Mode.} In this part, we investigate the impact of incorporating reasoning capabilities on model performance. We compare models trained with and without reasoning modes using the same training data, with results shown in Fig.~\ref{figure:combine_2.}(c). We find that omitting chain-of-thought leads to more efficient inference but results in a substantial performance drop. This clearly demonstrates the importance of incorporating reasoning abilities for verification tasks in scientific domains.
\section{Conclusion}
We highlight verification as a critical step toward advancing the scientific reasoning capabilities of LLMs. To this end, we introduce SCI-VerifyBench, a high-quality and diverse benchmark spanning mathematics, physics, chemistry, biology, and commonsense scientific QA tasks, designed to rigorously and systematically assess models’ cross-disciplinary scientific verification capabilities. Our study further demonstrates that chain-of-thought reasoning is essential for scientific verification, particularly when answers are complex or admit multiple equivalent forms. Building on this insight, we develop SCI-Verifier, a verifier endowed with concise reasoning abilities specifically tailored for verification tasks. Together, SCI-VerifyBench and SCI-Verifier provide both a comprehensive evaluation framework and a practical solution for scientific verification, offering strong potential to guide the continued advancement and reliability of LLMs in scientific reasoning.

\bibliography{iclr2026_conference}
\bibliographystyle{iclr2026_conference}

\clearpage

\appendix

\vspace{20pt}
\textbf{{\Large Appendix for SCI-Verifier}}

\section{Details of SCI-VerifyBench}\label{sec:appendix:statics}
In this section, we introduce details of the process of constructing SCI-VerifyBench, including the prompts and models used for data generation described in Sec.~\ref{sec:appendix:statics:generation}, the model annotation process and the prompts and parameters designed for practical use described in Sec.~\ref{sec:appendix:statics:eval}, as well as the data details and sample cases in SCI-VerifyBench described in Sec.~\ref{sec:appendix:statics:details}.
\subsection{Data Generation}\label{sec:appendix:statics:generation}
The data generation involves two parts. The first part uses LLMs to generate answers for existing questions, and correctness is determined by comparing the generated answers with the reference answers. The second part generates equivalent answers based on the characteristics of different subjects, testing whether the model can correctly recognize these equivalent forms. For both parts, multiple LLMs are used to generate candidate answers, including Qwen3-32B, Qwen3-30B-A3B-Thinking-2507, Qwen3-30B-A3B-Instruct-2507, LLaMa3.3-70B-Instruct, GPT-oss-20B, Qwen2.5-32B-Instruct, Gemma-3-27b-it, and Qwen3-8B. The prompt used in the first part is shown in Box.~\ref{a.1}. For the second part, different prompts are used for each subject according to the corresponding task. The Math prompts refer to Box.~\ref{a.2} to Box.~\ref{a.4}. The Physics prompts refer to Box.~\ref{p.1} to Box.~\ref{p.3}. The Chemistry prompts refer to Box.~\ref{chem.1} to Box.~\ref{chem.9}. The Biology prompts refer to Box.~\ref{bio.1} to Box.~\ref{bio.5}. The QA prompts refer to Box.~\ref{qa.1}.
\begin{center}
\begin{namedbox}[label=a.1]{Prompt for generating LLM response}
Please answer the problem adhering to the following rules:
1. Please use LaTeX format to represent the variables and formulas used in the solution process and results.
2. Please put the final answer(s) in boxed\{\}, note that the unit of the answer should not be included in boxed\{\}.
3. If the problem requires multiple answers, list them in order, each in a separate boxed\{\}.
Problem:\{{answer}\}
\end{namedbox}
\end{center}
\begin{center}
\begin{namedbox}[label=a.2]{Prompt for generating equivalent answers to mathematical interval problems}

You are given a mathematical interval answer.
Generate 10 different equivalent forms of this interval, using transformations such as:  

\begin{itemize}[noitemsep, topsep=0pt]
    \item Rewriting as inequalities: $\left[0,1\right] \rightarrow0 \leq x\leq 1$
    \item Rewriting as set operations:$\left[0,2\right] \cup \left[1,3\right] = \left[0,3\right]$
    \item Open/closed interval limit definitions: $\left(a,b\right) = \lim_{\epsilon\rightarrow0^+}\left[a+\epsilon, b - \epsilon\right]$
    \item Converting to numeric sets: $[0,2] \rightarrow \{{0,1,2}\} $ (for integer endpoints)
\end{itemize}

\textbf{Format}: Output exactly 10 forms. Each form must be wrapped in LaTeX \textbackslash boxed\{...\}. Separate each answer with a newline (\textbackslash n).

\textbf{Example}:

\hspace*{2mm}
Input: $\left[0,1\right]$

\hspace*{2mm}
Output:

\hspace*{4mm}
$\boxed{{{{0 \leq x \leq 1}}}}$

\hspace*{4mm}
$\boxed{{{{[0,1) \cup {{1}}}}}}$

\hspace*{4mm}
$\boxed{{{{(0-0.0001,1+0.0001)}}}}$

\hspace*{4mm}
$\boxed{{{{{{x \in \mathbb{{R}}}} \mid 0 \leq x \leq 1}}}}$

\hspace*{4mm}
$\boxed{{{{{{0,0.25,0.5,0.75,1}}}}}}$

Only output the converted results, do not output the conversion process!

Now, process the following answer: \{answer\}, 

\end{namedbox}
\end{center}

\begin{center}
\begin{namedbox}[label=a.3]{Prompt for generating equivalent answers to mathematical expression problems}
You are given a mathematical expression or a numeric answer.
Generate 10 different equivalent forms of this expression, using transformations such as:  

\begin{itemize}[noitemsep, topsep=0pt]
    \item Factoring or expansion: $x^2+2x+1 \rightarrow (x+1)^2$
    \item Fraction simplification: $(x^2 -1) / (x + 1) \rightarrow x - 1$
    \item Leaving fraction unsimplified: $(x^2 - 1) / (x+1)$ (unchanged)
    \item Partial fraction decomposition: $1 / (x(x+1)) \rightarrow 1/x - 1 / (x+1)$
    \item Fraction to decimal conversion: $1/2 \rightarrow 0.5$
    \item Trigonometric identities: $\sin^2x+\cos^2x=1$
    \item Trigonometric transformations: $\sin 2x = 2 \sin x \cos x$
    \item Taylor expansion: $\sin x \approx x - x^3/3!$
    \item Exponential/logarithm rules: $ln(ab)=ln a + ln b$
    \item Substitution: let y=x+1, then $x^2+2x+1 = y^2$
    \item Approximating special constants: $\pi \approx 3.14159$, $e \approx 2.718$
    \item Angle-radian conversion: $\pi /6 = 30^\circ$
\end{itemize}

\textbf{Format}: Output exactly 10 forms. Each form must be wrapped in LaTeX \textbackslash boxed\{...\}. Separate each answer with a newline (\textbackslash n). 

\textbf{Example}:

\hspace*{2mm}
Input: $(x+1)^2$

\hspace*{2mm}
Output:

\hspace*{4mm}
$\boxed{{{{(x+1)(x+1)}}}}$

\hspace*{4mm}
$\boxed{{{{x^2+2x+1}}}}$

\hspace*{4mm}
$\boxed{{{{\frac{{(x+1)^3}}{{x+1}}}}}}$

\hspace*{4mm}
$\boxed{{{{\text{{Let }} y=x+1, \\; y^2}}}}$

\hspace*{4mm}
$\boxed{{{{(x+1)^2 \approx 1 + 2x; \text{{ for small }} x}}}}$

Only output the converted results, do not output the conversion process!

Now, process the following answer: \{answer\},
\end{namedbox}
\end{center}

\begin{center}

\begin{namedbox}[label=a.4]{Prompt for generating equivalent answers to mathematical equation problems}
You are given a mathematical equation.
Generate 10 different equivalent forms of this equation, using transformations such as:  

\begin{itemize}[noitemsep, topsep=0pt]
    \item Moving terms: $x+3=5 \rightarrow x=2$
    \item Multiplying/dividing both sides: $2x=4 \rightarrow x=2$
    \item Factoring: $x^2-1=0 \rightarrow (x-1)(x+1)=0$
    \item Completing the square: $x^2+6x+5=0 \rightarrow (x+3)^2-4=0$
    \item Root extraction: $x^2=4 \rightarrow x=±2$
    \item Substitution in equations:$x^2+1=0 \rightarrow let x=i, then i^2+1=0$
    \item Trigonometric transformations: $\sin^2x=1-\cos^2x$
    \item Taylor expansion for approximate solution: $\sin x \approx x \rightarrow x \approx 0$
    \item Domain restrictions: $\sqrt{x-1}=x-3 \text{ requires } x\geq1$
\end{itemize}

\textbf{Format}: Output exactly 10 forms. Each form must be wrapped in LaTeX \textbackslash boxed\{...\}. Separate each answer with a newline (\textbackslash n).

\textbf{Example}:

\hspace*{2mm}
Input: 
$x^2 - 1 = 0$

\hspace*{2mm}
Output:

\hspace*{4mm}
$\boxed{{{{x^2=1}}}}$

\hspace*{4mm}
$\boxed{{{{(x-1)(x+1)=0}}}}$

\hspace*{4mm}
$\boxed{{{{x=\pm 1}}}}$

\hspace*{4mm}
$\boxed{{{{(x-0)^2 - 1=0}}}}$

\hspace*{4mm}
$\boxed{{{{\cos^2\theta - \sin^2\theta = 0 \quad \text{{(substitution }} x=\cos\theta)}}}}$

Only output the converted results, do not output the conversion process!

Now, process the following answer: \{answer\},
\end{namedbox}
\end{center}

\begin{center}
\begin{namedbox}[label=p.1]{Prompt for generating equivalent answers to physics numerical problems}

Given the following question and an answer in expression form, generate 10 different equivalent forms of this interval, using transformations such as:  

\begin{itemize}[noitemsep, topsep=0pt]
    \item Arithmetic operations or evaluation (e.g., $5+3 \rightarrow 8$)
    \item Substitution of variable values (e.g., $2x+3$,$x=2 \rightarrow 7$)
    \item Scientific notation, fractions, or decimals (e.g., $0.00012 \rightarrow 1.2×10^{-4}$)
    \item Unit conversion or dimensional adjustment (e.g., $1 km \rightarrow 1000 m$)
    \item Dimensional conversion (e.g., $1 N \rightarrow 1 kg\cdot m/s^2$)
\end{itemize}

\textbf{Format}: Output exactly 10 forms. Each form must be wrapped in LaTeX \textbackslash boxed\{...\}. Separate each answer with a newline (\textbackslash n).

\textbf{Example}:

\hspace*{2mm}
Input: [0,1]

\hspace*{2mm}
Output:

\hspace*{4mm}
$\boxed{{{{0 \leq x \leq 1}}}}$

\hspace*{4mm}
$\boxed{{{{[0,1) \cup {{1}}}}}}$

\hspace*{4mm}
$\boxed{{{{(0-0.0001,1+0.0001)}}}}$

\hspace*{4mm}
$\boxed{{{{{{x \in \mathbb{{R}}}} \mid 0 \leq x \leq 1}}}}$

\hspace*{4mm}
$\boxed{{{{{{0,0.25,0.5,0.75,1}}}}}}$

Only output the converted results, do not output the conversion process!

Now, process the following answer: 
Question: \{question\}.
Answer (Numeric): \{answer\}, 
\end{namedbox}
\end{center}

\begin{center}
\begin{namedbox}[label=p.2]{Prompt for generating equivalent answers to physics expression problems}
Given the following question and an answer in expression form, generate 10 different equivalent forms of this interval, using transformations such as: 

\begin{itemize}[noitemsep, topsep=0pt]
    \item  Factoring or expansion: $x^2+2x+1 \rightarrow (x+1)^2$
    \item Fraction simplification: $(x^2-1)/(x+1) \rightarrow x-1$
    \item Leaving fraction unsimplified: $(x^2-1)/(x+1)$ (unchanged)
    \item  Partial fraction decomposition: $1/(x(x+1)) \rightarrow 1/x - 1/(x+1)$
    \item  Fraction to decimal conversion:$1/2 \rightarrow 0.5$
    \item  Trigonometric identities: $\sin^2x+\cos^2x=1$
    \item  Trigonometric transformations: $\sin 2x = 2 \sin x \cos x$
    \item  Taylor expansion: $sin x \approx x - x^3/3!$
    \item  Exponential/logarithm rules: $ln(ab)=ln a + ln b$
    \item  Substitution: let $y=x+1\text{, then } x^2+2x+1 = y^2$
    \item  Approximating special constants: $\pi \approx 3.14159, e \approx 2.718$
    \item  Angle-radian conversion: $\pi/6 = 30°$
    \item  Dimensional conversion (e.g., $1 N \rightarrow 1 kg\cdot m/s^2$)
\end{itemize}

\textbf{Format}: Output exactly 10 forms. Each form must be wrapped in LaTeX \textbackslash boxed\{...\}. Separate each answer with a newline (\textbackslash n).

\textbf{Example}:

\hspace*{2mm}
Input: $(x+1)^2$

\hspace*{2mm}
Output:

\hspace*{4mm}
$\boxed{{{{(x+1)(x+1)}}}}$

\hspace*{4mm}
$\boxed{{{{x^2+2x+1}}}}$

\hspace*{4mm}
$\boxed{{{{\frac{{(x+1)^3}}{{x+1}}}}}}$

\hspace*{4mm}
$\boxed{{{{\text{{Let }} y=x+1; y^2}}}}$

\hspace*{4mm}
$\boxed{{{{(x+1)^2 \approx 1 + 2x \text{{; for small }} x}}}}$

Only output the converted results, do not output the conversion process!

Now, process the following answer: 
Question: \{question\}.
Answer (expression): \{answer\},

\end{namedbox}
\end{center}

\begin{center}

\begin{namedbox}[label=p.3]{Prompt for generating equivalent answers to physics equation problems}
Given the following question and an answer in expression form, generate 10 different equivalent forms of this interval, using transformations such as:

\begin{itemize}[noitemsep, topsep=0pt]
    \item Moving terms: $x+3=5 \rightarrow x=2$
    \item  Multiplying/dividing both sides: $2x=4 \rightarrow x=2$
    \item  Factoring: $x^2-1=0 \rightarrow (x-1)(x+1)=0$
    \item  Completing the square: $x^2+6x+5=0 \rightarrow (x+3)^2-4=0$
    \item  Root extraction: $x^2=4 \rightarrow x=\pm 2$
    \item  Substitution in equations: $x^2+1=0 \rightarrow let x=i, then i^2+1=0$
    \item  Trigonometric transformations: $\sin^2x=1-\cos^2x$
    \item Taylor expansion for approximate solution: $\sin x \approx x \rightarrow x \approx 0$
    \item  Domain restrictions: $\sqrt{(x - 1)}=x -3 \text{ requires } x\geq 1$
    \item  Dimensional conversion (e.g., $F = ma, m=1000 g, a=2 m/s² \rightarrow F = 2 N$)
\end{itemize}

\textbf{Format}: Output exactly 10 forms. Each form must be wrapped in LaTeX \textbackslash boxed\{...\}. Separate each answer with a newline (\textbackslash n).

\textbf{Example}:

\hspace*{2mm}
Input: $x^2 - 1 = 0$

\hspace*{2mm}
Output:

\hspace*{4mm}
$\boxed{{{{x^2=1}}}}$

\hspace*{4mm}
$\boxed{{{{(x-1)(x+1)=0}}}}$

\hspace*{4mm}
$\boxed{{{{x=\pm 1}}}}$

\hspace*{4mm}
$\boxed{{{{(x-0)^2 - 1=0}}}}$

\hspace*{4mm}
$\boxed{{{{\cos^2\theta - \sin^2\theta = 0 \quad \text{{(substitution }} x=\cos\theta)}}}}$

Only output the converted results, do not output the conversion process!

Now, process the following answer: 
Question: \{question\}.
Answer (equation): \{answer\}
\end{namedbox}
\end{center}

\begin{center}
\begin{namedbox}[label=chem.1]{Prompt for generating equivalent answers to chemical solvent prediction problems}

Given a chemical reaction and a predicted solvent, generate 10 equivalent forms of the solvent answer, considering:

\begin{itemize}[noitemsep, topsep=0pt]
    \item Different chemical names (IUPAC, common, or trivial)
    \item Abbreviations (e.g., EtOH)
    \item Molecular formulas (e.g., C2H5OH)
    \item SMILES representations
    \item Solvent class equivalence (e.g., polar protic, polar aprotic)
    \item Mixture equivalences (e.g., EtOH:H2O 1:1 $\equiv$ H2O:EtOH 1:1)
\end{itemize}

\textbf{Format}: Output exactly 10 forms, each wrapped in LaTeX \textbackslash boxed\{...\} and separated by newline.

\textbf{Example}:

\hspace*{2mm}
Input: Ethanol

\hspace*{2mm}
Output:

\hspace*{4mm}
$\boxed{{\text{{Ethanol}}}}$

\hspace*{4mm}
$\boxed{{{\text{EtOH}}}}$

\hspace*{4mm}
$\boxed{{{\text{C2H5OH}}}}$

\hspace*{4mm}
$\boxed{{{\text{CCO}}}}$

\hspace*{4mm}
$\boxed{{{\text{alcohol}}}}$

\hspace*{4mm}
$\boxed{{{\text{polar protic solvent}}}}$

\hspace*{4mm}
$\boxed{{{\text{EtOH:H2O 1:1}}}}$

\hspace*{4mm}
$\boxed{{{\text{H2O:EtOH 1:1}}}}$

\hspace*{4mm}
$\boxed{{{\text{ethyl alcohol}}}}$

\hspace*{4mm}
$\boxed{{\text{{C-C-O}}}}$

Only output the converted results, do not output the conversion process!

Now process:
Reaction: \{question\}
Predicted solvent: \{answer\},

\end{namedbox}
\end{center}

\begin{center}

\begin{namedbox}[label=chem.2]{Prompt for generating equivalent answers to chemical property prediction problems}
Given a molecule and a predicted property value, generate 10 equivalent forms, using:
\begin{itemize}[noitemsep, topsep=0pt]
    \item Unit conversions (e.g., Celsius $\leftrightarrow$ Kelvin $\leftrightarrow$ Fahrenheit)
    \item Ranges vs single values
    \item Approximate vs exact
    \item Different notations (e.g., logP, Kow)
    \item Descriptive labels (e.g., high solubility)
\end{itemize}

\textbf{Format}: 10 forms wrapped in LaTeX \textbackslash boxed\{...\}.

\textbf{Example}:

\hspace*{2mm}
Input: Water boiling point = 100 $^\circ$C

\hspace*{4mm}
Output:

\hspace*{4mm}
$\boxed{{{{100~^\circ \text{C}}}}}$

\hspace*{4mm}
$\boxed{{{{373~\text{K}}}}}$

\hspace*{4mm}
$\boxed{212 ^\circ \text{F}}$

\hspace*{4mm}
$\boxed{{{{100 - 101~^\circ \text{C}}}}}$

\hspace*{4mm}
$\boxed{{\text{{high boiling point}}}}$

\hspace*{4mm}
$\boxed{{{{\text{approx. } 100^\circ\text{C}}}}}$

\hspace*{4mm}
$\boxed{{{{373.15 \text{K}}}}}$

\hspace*{4mm}
$\boxed{{{{0.1\times10^3~^\circ\text{C}}}}}$

\hspace*{4mm}
$\boxed{{{{100\text{ Celsius}}}}}$

\hspace*{4mm}
$\boxed{\text{water boils at 100}^\circ\text{C}}$

Only output the converted results, do not output the conversion process!
Now process:
Molecule: \{question\}
Property: \{answer\},

\end{namedbox}
\end{center}

\begin{center}

\begin{namedbox}[label=chem.3]{Prompt for generating equivalent answers to chemical yield prediction problems}
Given a chemical reaction and a predicted yield, generate 10 equivalent forms:
\begin{itemize}[noitemsep, topsep=0pt]
    \item  Percentage $\leftrightarrow$ decimal
    \item  Ranges $\leftrightarrow$ single value
    \item  Descriptive labels (e.g., high yield, trace)
    \item  Approximate expressions
\end{itemize}

\textbf{Format}: 10 forms, LaTeX \textbackslash boxed\{...\}, newline separated.

\textbf{Example}:

\hspace*{2mm}
Input: 85\%

\hspace*{2mm}
Output:

\hspace*{4mm}
$\boxed{{{{85\%}}}}$

\hspace*{4mm}
$\boxed{{{{0.85}}}}$

\hspace*{4mm}
$\boxed{\text{{{high yield}}}}$

\hspace*{4mm}
$\boxed{{{{80-90\%}}}}$

\hspace*{4mm}
$\boxed{{{{\text{approx. } 85\%}}}}$

\hspace*{4mm}
$\boxed{{{{y = 0.85}}}}$

\hspace*{4mm}
$\boxed{{{{\text{yield} > 80\%}}}}$

\hspace*{4mm}
$\boxed{{{\text{quantitative yield}}}}$

\hspace*{4mm}
$\boxed{{{\text{major product}}}}$

\hspace*{4mm}
$\boxed{{{\text{trace product}}}}$

Only output the converted results, do not output the conversion process!

Now process:
Reaction: \{question\}
Predicted yield: \{answer\},

\end{namedbox}
\end{center}

\begin{center}

\begin{namedbox}[label=chem.4]{ Prompt for generating equivalent answers to chemical retrosynthesis problems}

Given a target molecule, generate 10 equivalent retrosynthesis answers:
\begin{itemize}[noitemsep, topsep=0pt]
    \item  Different valid disconnections
    \item  Different reagents or reducing / oxidizing agents
    \item  Alternative synthetic routes (single-step or multi-step)
    \item  Protecting group alternatives
    \item  Functional group or stereochemical equivalence
\end{itemize}

\textbf{Format}: 10 forms wrapped in LaTeX \textbackslash boxed\{...\}.

\textbf{Example}:

\hspace*{2mm}
Input: Target = Benzyl alcohol

\hspace*{2mm}
Output:

\hspace*{4mm}
$\boxed{{{{\text{Benzaldehyde + NaBH4 }\rightarrow \text{Benzyl alcohol}}}}}$

\hspace*{4mm}
$\boxed{{{{\text{Benzyl chloride + NaOH} \rightarrow \text{Benzyl alcohol}}}}}$

\hspace*{4mm}
$\boxed{{{\text{Benzaldehyde reduced by LiAlH4}}}}$

\hspace*{4mm}
$\boxed{{{{\text{Benzyl bromide + KOH }\rightarrow \text{Benzyl alcohol}}}}}$

\hspace*{4mm}
$\boxed{{{{\text{Benzaldehyde } \rightarrow \text{H2 + Pd/C → Benzyl alcohol}}}}}$

\hspace*{4mm}
$\boxed{{{{\text{C6H5CH2Cl + NaOH }\rightarrow \text{C6H5CH2OH}}}}}$

\hspace*{4mm}
$\boxed{{\text{{Alternative protecting group: Boc route}}}}$

\hspace*{4mm}
$\boxed{{{\text{Multi-step oxidation-reduction route}}}}$

\hspace*{4mm}
$\boxed{{{\text{Reductive amination route to same alcohol}}}}$

\hspace*{4mm}
$\boxed{{{\text{Electrochemical reduction route}}}}$

Only output the converted results, do not output the conversion process!

Now process:
Target molecule: \{question\}
Retrosynthesis answer: \{answer\},

\end{namedbox}
\end{center}

\begin{center}
\begin{namedbox}[label=chem.5]{ Prompt for generating equivalent answers to chemical temperature prediction problems}
Given a reaction and predicted temperature, generate 10 equivalent forms:
\begin{itemize}[noitemsep, topsep=0pt]
    \item  Different units ($^\circ$C, K, $^\circ$F)
    \item  Ranges vs single value
    \item  Descriptive labels (low, room temp, high)
    \item  Common lab descriptions (ice bath, reflux)
\end{itemize}

\textbf{Format}: LaTeX \textbackslash boxed\{...\}, 10 forms.

\textbf{Example}:

\hspace*{2mm}
Input: 80$^\circ$C

\hspace*{2mm}
Output:

\hspace*{4mm}
$\boxed{{{{80~^\circ \text{C}}}}}$

\hspace*{4mm}
$\boxed{{{{353~\text{K}}}}}$

\hspace*{4mm}
$\boxed{{{{176~^\circ \text{F}}}}}$

\hspace*{4mm}
$\boxed{{{{70-90~^\circ \text{C}}}}}$

\hspace*{4mm}
$\boxed{{{{\text{approx. }80^\circ\text{C}}}}}$

\hspace*{4mm}
$\boxed{{{\text{reflux}}}}$

\hspace*{4mm}
$\boxed{{{\text{room temperature}}}}$

\hspace*{4mm}
$\boxed{{{\text{ice bath}}}}$

\hspace*{4mm}
$\boxed{{{\text{moderate heating}}}}$

\hspace*{4mm}
$\boxed{\text{{{high temperature}}}}$

Only output the converted results, do not output the conversion process!

Now process:
Reaction: \{question\}
Predicted temperature: \{answer\},
\end{namedbox}
\end{center}

\begin{center}

\begin{namedbox}[label=chem.6]{Prompt for generating equivalent answers to chemical product prediction problems}
Given reactants and reaction conditions, generate 10 equivalent forms of the predicted product:
\begin{itemize}[noitemsep, topsep=0pt]
    \item  Different structure representations: SMILES, InChI, molecular formula
    \item  IUPAC name, common name, trivial name
    \item  Stereoisomers (R/S)
    \item  Tautomers, salts, hydrates
    \item  Alternative valid representations (chair/boat conformers)
\end{itemize}

\textbf{Format}: 10 LaTeX \textbackslash boxed\{...\} outputs.

\textbf{Example}:

\hspace*{2mm}
Input: Nitrobenzene

\hspace*{2mm}
Output:

\hspace*{4mm}
$\boxed{{{\text{C6H5NO2}}}}$

\hspace*{4mm}
$\boxed{{{\text{c1ccc(cc1)[N+](=O)[O-]}}}}$

\hspace*{4mm}
$\boxed{{\text{{Nitrobenzene}}}}$

\hspace*{4mm}
$\boxed{{{\text{Benzene nitro compound}}}}$

\hspace*{4mm}
$\boxed{{\text{{C6H5-NO2}}}}$

\hspace*{4mm}
$\boxed{\text{{{Nitrobenzol}}}}$

\hspace*{4mm}
$\boxed{{{\text{C6H5NO2} \cdot \text{H2O}}}}$

\hspace*{4mm}
$\boxed{{{\text{aromatic nitro compound}}}}$

\hspace*{4mm}
$\boxed{{\text{{benzene derivative}}}}$

\hspace*{4mm}
$\boxed{{\text{{Racemic mixture if applicable}}}}$

Only output the converted results, do not output the conversion process!

Now process:
Reactants: \{question\}
Predicted product: \{answer\},
\end{namedbox}
\end{center}

\begin{center}
\begin{namedbox}[label=chem.7]{Prompt for generating equivalent answers to chemical mol2caption problems}
Given a molecular structure, generate 10 equivalent textual descriptions:
\begin{itemize}[noitemsep, topsep=0pt]
    \item IUPAC name
    \item  Common/trivial names
    \item  Chinese name
    \item  Functional description (solvent, reagent, drug)
    \item  Usage or property description
\end{itemize}

\textbf{Format}: LaTeX \textbackslash boxed\{...\}, 10 forms.

\textbf{Example}:

\hspace*{2mm}
Input: C2H5OH

\hspace*{2mm}
{Output}:

\hspace*{4mm}
$\boxed{{\text{{Ethanol}}}}$

\hspace*{4mm}
$\boxed{\text{{{EtOH}}}}$

\hspace*{4mm}
$\boxed{{{\text{alcohol}}}}$


\hspace*{4mm}
$\boxed{{\text{{ethyl alcohol}}}}$

\hspace*{4mm}
$\boxed{{\text{{common solvent}}}}$

\hspace*{4mm}
$\boxed{{\text{{disinfectant}}}}$

\hspace*{4mm}
$\boxed{{{\text{flammable liquid}}}}$

\hspace*{4mm}
$\boxed{{\text{{reagent in reactions}}}}$

\hspace*{4mm}
$\boxed{{{\text{soluble in water}}}}$

Only output the converted results, do not output the conversion process!

Now process:
Molecule: \{question\}
Mol2caption answer: \{answer\},
\end{namedbox}
\end{center}

\begin{center}

\begin{namedbox}[label=chem.8]{Prompt for generating equivalent answers to chemical caption2mol problems}
Given a textual description of a molecule, generate 10 equivalent molecular representations:
\begin{itemize}[noitemsep, topsep=0pt]
    \item Different SMILES strings (canonical and non-canonical)
    \item  InChI
    \item  Molecular formula
    \item  Graphical structure representation (if feasible)
    \item  Different naming conventions resolved to the same structure
\end{itemize}

Format: LaTeX \textbackslash boxed\{...\}, 10 forms.

\textbf{Example}:
\hspace*{2mm}
Input: "alcohol used for disinfection"

\hspace*{2mm}
Output:

\hspace*{4mm}
$\boxed{{{\text{C2H5OH}}}}$

\hspace*{4mm}
$\boxed{{\text{{CCO}}}}$

\hspace*{4mm}
$\boxed{{\text{{OCC}}}}$

\hspace*{4mm}
$\boxed{{\text{{InChI=1S/C2H6O/c1-2-3/h3H,2H2,1H3}}}}$

\hspace*{4mm}
$\boxed{{\text{{ethanol}}}}$

\hspace*{4mm}
$\boxed{\text{{{EtOH}}}}$

\hspace*{4mm}
$\boxed{\text{{{ethyl alcohol}}}}$

\hspace*{4mm}
$\boxed{\text{{{common lab alcohol}}}}$

\hspace*{4mm}
$\boxed{{{\text{solvent for reactions}}}}$

Only output the converted results, do not output the conversion process!

Now process:
Caption: {question}
Predicted molecule: {answer},
\end{namedbox}
\end{center}

\begin{center}

\begin{namedbox}[label=chem.9]{Prompt for generating equivalent answers to chemical name conversion problems}
Given a molecule name in one format, generate 10 equivalent forms:
\begin{itemize}[noitemsep, topsep=0pt]
    \item  IUPAC $\leftrightarrow$ trivial/common name
    \item  English $\leftrightarrow$ Chinese name
    \item  SMILES
    \item  InChI
    \item  Hydrate/salt forms if applicable
    \item  Synonyms and international spelling variants
\end{itemize}

\textbf{Format}: LaTeX \textbackslash boxed\{...\}, 10 outputs.

\textbf{Example}:

\hspace*{2mm}
Input: Acetic acid

\hspace*{2mm}
Output:

\hspace*{4mm}
$\boxed{{\text{{Acetic acid}}}}$

\hspace*{4mm}
$\boxed{{\text{{ethanoic acid}}}}$

\hspace*{4mm}
$\boxed{{{\text{CH3COOH}}}}$

\hspace*{4mm}
$\boxed{{{\text{C2H4O2}}}}$

\hspace*{4mm}
$\boxed{\text{{{Vinegar acid}}}}$

\hspace*{4mm}
$\boxed{\text{{{Acetic acid, glacial}}}}$

\hspace*{4mm}
$\boxed{{{\text{Acetic acid aqueous solution}}}}$

\hspace*{4mm}
$\boxed{{\text{{Sodium acetate form}}}}$

\hspace*{4mm}
$\boxed{\text{{{Acid CH3COOH}}}}$

Only output the converted results, do not output the conversion process!

Now process:
Name: \{question\}
Predicted conversion: \{answer\}
\end{namedbox}
\end{center}

\begin{center}
\begin{namedbox}[label=bio.1]{Prompt for generating equivalent answers to biological protein inverse folding problems}
Given the following question and an answer in protein sequence form, generate 10 different equivalent forms of this sequence, using transformations such as: 

\begin{itemize}[noitemsep, topsep=0pt]
    \item Synonymous sequences folding into the same structure
    \item - Conservative amino acid substitutions (e.g., Lys $\leftrightarrow$ Arg, Asp $\leftrightarrow$ Glu)
    \item FASTA format $\leftrightarrow$ plain sequence string
    \item One-letter code $\leftrightarrow$ three-letter code
    \item Sequence with small tolerated mutations that preserve structure
    \item Adding/removing header lines in FASTA
    \item Using lowercase vs uppercase amino acid letters
    \item Introducing gap symbols to indicate alignment but preserving structure
    \item Representing sequence in JSON or array form
    \item Grouping amino acids by domains/regions but preserving the overall sequence
\end{itemize}

\textbf{Format}: Output exactly 10 forms. Each form must be wrapped in LaTeX \textbackslash boxed\{...\}. 

\textbf{Example}:  

\hspace*{2mm}
Input: MKAILVVLLYTAA  

\hspace*{2mm}
Output:  

\hspace*{4mm}
$\boxed{\text{{{MKAILVVLLYTAA}}}}$

\hspace*{4mm}
$\boxed{{{\text{Met-Lys-Ala-Ile-Leu-Val-Val-Leu-Leu-Tyr-Thr-Ala-Ala}}}}  $

\hspace*{4mm}
$\boxed{{\text{{>seq1\\nMKAILVVLLYTAA}}}}  $

\hspace*{4mm}
$\boxed{{\text{{mkailvvll ytaa}}}}  $

\hspace*{4mm}
$\boxed{{{\text{MKAILVVL--YTAA}}}} $

\hspace*{4mm}
$\boxed{{\text{{[`M',`K',`A',`I',`L',`V',`V',`L',`L',`Y',`T',`A',`A']}}}}  $

\hspace*{4mm}
$\boxed{{{\text{MKAILVVLLY(S/T)AA}}}}  $

\hspace*{4mm}
$\boxed{{\text{{MKAILVVLLY RAA (Arg substitution)}}}}$

\hspace*{4mm}
$\boxed{{{\text{[Domain1: MKAILV, Domain2: VLLYTAA]}}}} $ 

\hspace*{4mm}
$\boxed{{\text{{MKAILVVLLYTAA (unchanged)}}}} $

Only output the converted results in LaTeX \\boxed{{{{...}}}},  do not output the conversion process!

Now, process the following answer:  
Question: \{question\}  
Answer: \{answer\},
\end{namedbox}
\end{center}

\begin{center}

\begin{namedbox}[label=bio.2]{Prompt for generating equivalent answers to biological protein structure prediction problems}
Given the following question and an answer in protein structure form, generate 10 different equivalent forms of this structure, using transformations such as: 

\begin{itemize}[noitemsep, topsep=0pt]
    \item PDB format $\leftrightarrow$ mmCIF format
    \item  3D atomic coordinates $\leftrightarrow$ C$\alpha$-only coordinates
    \item  3D structure $\leftrightarrow$ contact map $\leftrightarrow$ distance matrix
    \item  Cartesian coordinates $\leftrightarrow$ internal torsion angles ($\phi$, $\psi$, $\chi$)
    \item Secondary structure representation (helix/sheet/loop) instead of full 3D
    \item  Alternative but equivalent chain numbering or atom ordering
    \item  Superimposed structures with RMSD < threshold
    \item  JSON or graph-based adjacency representation of the structure
    \item  Coarse-grained models (e.g., backbone only)
    \item  Visual encodings (e.g., dot-bracket for secondary structure, schematic diagrams)
\end{itemize}

\textbf{Format}: Output exactly 10 forms. Each form must be wrapped in LaTeX \textbackslash boxed\{...\}. 

\textbf{Example}:  

\hspace*{2mm}
Input: PDB file with C$\alpha$ coordinates of a helix  

\hspace*{2mm}
Output:  

\hspace*{4mm}
$\boxed{{\text{{ATOM 1 CA ALA 1 0.000 0.000 0.000}}}}  $

\hspace*{4mm}
$\boxed{\text{{{loop helix helix sheet}}}}  $

\hspace*{4mm}
$\boxed{\text{{{contact map: (1,5),(2,6),(3,7)}}}} $

\hspace*{4mm}
$\boxed{{{\text{distance matrix: [[0,1.2,...],[1.2,0,...],...]}}}}  $

\hspace*{4mm}
$\boxed{{{{\text{torsion angles } \phi=-60, \psi=-45}}}} $

\hspace*{4mm}
$\boxed{{{\text{mmCIF equivalent entry}}}}  $

\hspace*{4mm}
$\boxed{{\text{{JSON: {{"atoms":[{{"res":"ALA","atom":"CA","x":0,"y":0,"z":0}}]}}}}}}  $

\hspace*{4mm}
$\boxed{{{\text{graph: nodes=7, edges=[(1,5),(2,6),(3,7)]}}}}  $

\hspace*{4mm}
$\boxed{{{\text{backbone-only representation}}}} $

\hspace*{4mm}
$\boxed{{\text{{cartoon schematic: helix symbol}}}}  $

Only output the converted results in LaTeX \textbackslash boxed\{...\},  do not output the conversion process!

Now, process the following answer:  
Question: \{question\}  
Answer: \{answer\},
\end{namedbox}
\end{center}

\begin{center}
\begin{namedbox}[label=bio.3]{Prompt for generating equivalent answers to biological transformed agentclinic problems}
Given the following question and an answer in agent action/trajectory form, generate 10 different equivalent forms of this solution, using transformations such as: 

\begin{itemize}[noitemsep, topsep=0pt]
    \item  Different but equivalent action sequences leading to same outcome
    \item  Merging multiple actions into macro-actions
    \item  Splitting macro-actions into finer-grained steps
    \item  JSON $\leftrightarrow$ tabular $\leftrightarrow$ natural language action logs
    \item  Reordering independent actions without changing outcome
    \item  Adding no-op (null) actions that do not affect outcome
    \item  Representing state transitions instead of actions
    \item  Graph or tree representation of the policy trace
    \item  Summarizing actions at high-level clinical outcome description
    \item Encoding with symbolic tokens instead of natural language
\end{itemize}

\textbf{Format}: Output exactly 10 forms. Each form must be wrapped in LaTeX \textbackslash boxed\{...\}. 

\textbf{Example}:  

\hspace*{2mm}
Input: Give drug A $\rightarrow$ Measure blood pressure $\rightarrow$  Stop treatment  

\hspace*{2mm}
Output:  

\hspace*{4mm}
$\boxed{{{\text{Give drug A → Measure BP → Stop treatment}}}}  $

\hspace*{4mm}
$\boxed{\text{{{Action1, Action2, Action3}}}} $

\hspace*{4mm}
$\boxed{{\text{{[Give drug A, Measure blood pressure, Stop treatment]}}}}  $

\hspace*{4mm}
$\boxed{\text{{{MacroAction: Treat+Monitor}}}}  $

\hspace*{4mm}
$\boxed{{{{\text{State transitions: S0}\rightarrow \text{S1}\rightarrow\text{S2}}}}}  $

\hspace*{4mm}
$\boxed{{{{\text{Add no-op: Give drug A }\rightarrow \text{Wait} \rightarrow \text{Measure BP} \rightarrow \text{Stop}}}}}  $

\hspace*{4mm}
$\boxed{\text{{{JSON: {{``actions":[``Give drug A",``Measure BP",``Stop"]}}}}}}  $

\hspace*{4mm}
$\boxed{\text{{{Table: 1. Give drug A | 2. Measure BP | 3. Stop}}}}  $

\hspace*{4mm}
$\boxed{{{\text{Graph representation: nodes=states, edges=actions}}}} $ 

\hspace*{4mm}
$\boxed{{{\text{Outcome: Patient stabilized after drug A}}}}  $

Only output the converted results in LaTeX \textbackslash boxed\{...\}, do not output the conversion process!

Now, process the following answer: 
Question: {question}  
Answer: {answer}""",
\end{namedbox}
\end{center}

\begin{center}

\begin{namedbox}[label=bio.4]{Prompt for generating equivalent answers to biological rna structure prediction problems}
Given the following question and an answer in RNA structure form, generate 10 different equivalent forms of this structure, using transformations such as: 

\begin{itemize}[noitemsep, topsep=0pt]
    \item  Dot-bracket notation $\leftrightarrow$ CT format $\leftrightarrow$ BPseq format
    \item  Base-pairing list $\leftrightarrow$ adjacency matrix representation
    \item  2D secondary structure $\leftrightarrow$ 3D atomic coordinates
    \item  Minimal free energy structure $\leftrightarrow$ near-optimal structure
    \item Adding base-pair probabilities annotation
    \item Grouping stems/loops/bulges into modular blocks
    \item  JSON array representation of base pairs
    \item ASCII art of RNA fold representation
    \item Graph representation of paired/unpaired nodes
    \item - Annotated sequence with paired regions in brackets
\end{itemize}

\textbf{Format}: Output exactly 10 forms. Each form must be wrapped in LaTeX \textbackslash boxed\{...\}. 

\textbf{Example}:  

\hspace*{2mm}
Input: GCGCUUAGC, Structure: (((...)))  

\hspace*{2mm}
Output:  

\hspace*{4mm}
$\boxed{{\text{{(((...)))}}}} $

\hspace*{4mm}
$\boxed{{\text{{dot-bracket: (((...)))}}}} $

\hspace*{4mm}
$\boxed{{\text{{CT: 1 9, 2 8, 3 7}}}}  $

\hspace*{4mm}
$\boxed{{\text{{BPseq: 1 9, 2 8, 3 7}}}}  $

\hspace*{4mm}
$\boxed{{{\text{pairs=[(1,9),(2,8),(3,7)]}}}} $

\hspace*{4mm}
$\boxed{\text{{{matrix: [[0,0,1,...],...]}}}} $

\hspace*{4mm}
$\boxed{{{\text{near-optimal structure (((..).))}}}} $

\hspace*{4mm}
$\boxed{{\text{{ASCII: stem-loop diagram}}}}  $

\hspace*{4mm}
$\boxed{\text{{{JSON: {{``pairs":[[1,9],[2,8],[3,7]]}}}}}} $

\hspace*{4mm}
$\boxed{{{\text{annotated sequence: GCG(CUU)AGC}}}}  $

Only output the converted results in LaTeX \textbackslash boxed\{...\},  do not output the conversion process!

Now, process the following answer:  
Question: \{question\}  
Answer: \{answer\}
\end{namedbox}
\end{center}

\begin{center}

\begin{namedbox}[label=bio.5]{Prompt for generating equivalent answers to biological rna inverse folding problems}
Given the following question and an answer in RNA sequence form, generate 10 different equivalent forms of this sequence, using transformations such as: 

\begin{itemize}[noitemsep, topsep=0pt]
    \item Different sequences folding into the same target structure
    \item  U $\leftrightarrow$ T replacement (RNA vs DNA notation)
    \item  FASTA format $\leftrightarrow$ plain string sequence
    \item  Lowercase vs uppercase bases
    \item  Adding alignment gaps without changing fold
    \item  JSON or array encoding of sequence
    \item  Annotating sequence with regions (stem, loop, bulge)
    \item  Replacing synonymous positions with alternative nucleotides that preserve folding
    \item  Annotated sequence with secondary structure alignment marks
    \item  Splitting long sequence into chunks and recombining
\end{itemize}

\textbf{Format}: Output exactly 10 forms. Each form must be wrapped in LaTeX \textbackslash boxed\{...\}. 

\textbf{Example}:  
\hspace*{2mm}
Input: AUGCGAU  

\hspace*{2mm}
Output:  

\hspace*{4mm}
$\boxed{\text{{{AUGCGAU}}}}  $

\hspace*{4mm}
$\boxed{\text{{{augcgau}}}}  $

\hspace*{4mm}
$\boxed{{{\text{ATGCGAT}}}}  $

\hspace*{4mm}
$\boxed{{\text{{>seq1\\nAUGCGAU}}}}  $

\hspace*{4mm}
$\boxed{{\text{{AU-GCGAU}}}}  $

\hspace*{4mm}
$\boxed{{{\text{['A','U','G','C','G','A','U']}}}}  $

\hspace*{4mm}
$\boxed{\text{{{stem(AUG) loop(CGAU)}}}}  $

\hspace*{4mm}
$\boxed{{\text{{AUGCGAU (unchanged)}}}}  $

\hspace*{4mm}
$\boxed{{\text{{JSON: {{"sequence":"AUGCGAU"}}}}}} $ 

\hspace*{4mm}
$\boxed{\text{{{chunks: AU | GC | GA | U}}}}  $

Only output the converted results in LaTeX \textbackslash boxed\{...\},  do not output the conversion process!

Now, process the following answer:  
Question: \{question\}  
Answer: \{answer\}
\end{namedbox}
\end{center}

\begin{center}
\begin{namedbox}[label=qa.1]{Prompt for generating equivalent answers to QA problems}
You are given an answer to a general question.
Generate 10 different equivalent forms of this answer, using transformations such as:

\begin{itemize}[noitemsep, topsep=0pt]
    \item Paraphrasing the text while keeping the meaning the same.
    \item Changing the sentence structure or word order.
    \item Using synonyms or alternative expressions.
    \item Expressing the answer as a list, table, or bullet points if applicable.
    \item Rewriting as formal statements, equations, or logical expressions if relevant.
    \item Approximating special constants: $\pi \approx 3.14159, e \approx 2.718$
    \item Angle-radian conversion: $\pi/6 = 30^\circ$
    \item Dimensional conversion (e.g., $1 N \rightarrow 1 kg\cdot m/s^2$)
\end{itemize}

\textbf{Format}: Output exactly 10 forms. Each form must be wrapped in LaTeX \textbackslash boxed\{...\}. Separate each answer with a newline (\textbackslash n).

\textbf{Example}:

\hspace*{2mm}
Input: "Water boils at 100 degrees Celsius at sea level."

\hspace*{2mm}
{Output}:

\hspace*{4mm}
$\boxed{{\text{{Water reaches its boiling point at 100°C at sea level.}}}}$

\hspace*{4mm}
$\boxed{{{\text{At sea level, the boiling temperature of water is 100 degrees Celsius.}}}}$

\hspace*{4mm}
$\boxed{{{\text{100°C is the temperature at which water boils at sea level.}}}}$

Only output the converted results, do not output the conversion process!

Now, process the following answer: 
Question: \{question\}
Answer (expression): \{answer\}
\end{namedbox}
\end{center}

\subsection{Data Annotation And Evaluation}\label{sec:appendix:statics:eval}
Next, we describe the configurations and prompts used during data annotation and the actual evaluation. In this process, the inputs are the question, the reference answer, and the answer to be evaluated, and the output is the correctness of the answer being evaluated. To ensure stable outputs during the experiments, a temperature of 0 is used. The prompts with CoT are shown in Box.~\ref{inference_cot}, the prompts without CoT are shown in Box.~\ref{no_cot_inference}, and the prompts used in the main experiments to measure prompt stability are shown in Box.~\ref{other_inference}. The LLMs used during the data annotation process were Qwen3-30B-A3B-Instruct-2507, GPT-oss-20B, Qwen2.5-72B-Instruct, LLaMa3.3-Instruct, and CompassVerifier-32B. 
\clearpage

\begin{center}
\begin{namedbox}[width=0.9\linewidth, fontupper=\small,label=inference_cot]{Prompt of Inference with Thinking}
As a grading reward model, your task is to evaluate whether the candidate's final answer matches the provided standard answer. 
You must first output a detailed step-by-step analysis, then give a final structured judgment. 
Do not regenerate or improve answers, only compare.

\textbf{Evaluation Protocol:}

\textbf{1. Reference Standard:}
\begin{itemize}[noitemsep, topsep=0pt]
    \item The standard (gold) answer is definitive and always correct.
    \item The question is always valid — never challenge it.
    \item Do not regenerate answers; only compare candidate's answer with the gold answer.
\end{itemize}

\textbf{2. Comparison Method:}
\begin{itemize}[noitemsep, topsep=0pt]
    \item Analyze the question's requirements and the gold answer's structure.
    \item Determine if the question requires exact matching or allows equivalence.
    \item Compare ONLY the candidate's final answer. Ignore reasoning errors.
    \item Ignore differences in formatting or style.
    \item For math expressions: check algebraic equivalence step by step; if uncertain, test numerically at multiple points.
    \item For multiple-choice: only compare the final choice and its content.
\end{itemize}

\textbf{3. Multi-part Answers:}
\begin{itemize}[noitemsep, topsep=0pt]
    \item All parts must match the gold answer exactly.
    \item Partial matches are incorrect.
    \item If not specified, order may vary. For example, $\frac{27}{7}, -\frac{8}{7}$ and $-\frac{8}{7}, \frac{27}{7}$ are equivalent.
\end{itemize}

\textbf{4. Validity Check:}
\begin{itemize}[noitemsep, topsep=0pt]
    \item If incomplete (cut off, unfinished sentence) → Label as \textbf{INCOMPLETE}.
    \item If repetitive (looping words/phrases) → Label as \textbf{REPETITIVE}.
    \item If explicit refusal (e.g., "I cannot answer...") → Label as \textbf{REFUSAL}.
    \item Gives an answer but then negates it at the end → Label as \textbf{REFUSAL}.
    \item Any of the above → classify as \textbf{C} with the correct error type.
\end{itemize}

\textbf{Grading Scale:}

\textbf{\boxed{A} - CORRECT:}
\begin{itemize}[noitemsep, topsep=0pt]
    \item Matches gold exactly or equivalent (including algebraic/numeric equivalence).
    \item For numerical values: equivalent if equal under rounding tolerance.
    \item Semantic equivalence allowed.
\end{itemize}

\textbf{\boxed{B} - INCORRECT:}
\begin{itemize}[noitemsep, topsep=0pt]
    \item Any deviation from gold.
    \item Partial matches for multi-part answers.
\end{itemize}

\textbf{\boxed{C} - INCOMPLETE/REPETITIVE/REFUSAL:}
\begin{itemize}[noitemsep, topsep=0pt]
    \item Invalid answers (must specify error type).
\end{itemize}

\textbf{Execution Steps and Output:}

\textbf{Analysis step by step:}
\begin{enumerate}[noitemsep, topsep=0pt]
    \item First check validity (INCOMPLETE, REPETITIVE, REFUSAL). 
    \item Compare candidate’s final answer vs. gold answer in detail. 
        \begin{itemize}[noitemsep, topsep=0pt]
            \item Identify strict requirements (e.g., exact match, order, completeness).
            \item Allow tolerances (format differences, equivalent math forms, unsimplified fractions, provide the full answer for completion-type questions). 
            \item Check for equivalences, e.g., 
            $\frac{2x-7}{(x+1)(x-2)} \text{ and } \frac{3}{x+1} - \frac{1}{x-2}$\ are equivalent.
            
            \item Consider:
            \begin{itemize}
                \item Factoring/expansion: $x^2+2x+1 \to (x+1)^2$
                \item Fraction simplification: $\frac{x^2-1}{x+1} \to x-1$
                \item Partial fraction decomposition, decimals, trig identities, substitutions, etc.
            \end{itemize}
            \item For multiple-choice, answer must exactly match gold or be fully equivalent.
        \end{itemize}
    \item Provide thorough reasoning chain, highlighting subtle equivalences or deviations.
\end{enumerate}

\textbf{Final Judgment:} A/B/C

Here is your task:

\textless Original Question Begin\textgreater
\{question\}
\textless Original Question End\textgreater

\textless Standard Answer Begin\textgreater
\{gold answer\}
\textless Standard Answer End\textgreater

\textless Candidate's Answer Begin\textgreater
\{llm response\}
\textless Candidate's Answer End\textgreater

Analysis step by step (not to try solving the problem yourself) and Final Judgment:

\end{namedbox}
\end{center}
\begin{center}
\begin{namedbox}[width=0.9\linewidth, fontupper=\small,label=no_cot_inference]{Prompt of Inference without Thinking}
As a grading reward model, your task is to evaluate whether the candidate's final answer matches the provided standard answer. 
You must only give a final structured judgment. 
Do not regenerate or improve answers, only compare.

\textbf{Evaluation Protocol:}

\textbf{1. Reference Standard:}
\begin{itemize}[noitemsep, topsep=0pt]
    \item The standard (gold) answer is definitive and always correct.
    \item The question is always valid — never challenge it.
    \item Do not regenerate answers; only compare candidate's answer with the gold answer.
\end{itemize}

\textbf{2. Comparison Method:}
\begin{itemize}[noitemsep, topsep=0pt]
    \item Analyze the question's requirements and the gold answer's structure.
    \item Determine if the question requires exact matching or allows equivalence.
    \item Compare ONLY the candidate's final answer. Ignore reasoning errors.
    \item Ignore differences in formatting or style.
    \item For math expressions: check algebraic equivalence step by step; if uncertain, test numerically at multiple points.
    \item For multiple-choice: only compare the final choice and its content.
\end{itemize}

\textbf{3. Multi-part Answers:}
\begin{itemize}[noitemsep, topsep=0pt]
    \item All parts must match the gold answer exactly.
    \item Partial matches are incorrect.
    \item If not specified, order may vary. For example, $\frac{27}{7}, -\frac{8}{7}$ and $-\frac{8}{7}, \frac{27}{7}$ are equivalent.
\end{itemize}

\textbf{4. Validity Check:}
\begin{itemize}[noitemsep, topsep=0pt]
    \item If incomplete (cut off, unfinished sentence) → Label as \textbf{INCOMPLETE}.
    \item If repetitive (looping words/phrases) → Label as \textbf{REPETITIVE}.
    \item If explicit refusal (e.g., "I cannot answer...") → Label as \textbf{REFUSAL}.
    \item Gives an answer but then negates it at the end → Label as \textbf{REFUSAL}.
    \item Any of the above → classify as \textbf{C} with the correct error type.
\end{itemize}

\textbf{Grading Scale:}

\textbf{\boxed{A} - CORRECT:}
\begin{itemize}[noitemsep, topsep=0pt]
    \item Matches gold exactly or equivalent (including algebraic/numeric equivalence).
    \item For numerical values: equivalent if equal under rounding tolerance.
    \item Semantic equivalence allowed.
\end{itemize}

\textbf{\boxed{B} - INCORRECT:}
\begin{itemize}[noitemsep, topsep=0pt]
    \item Any deviation from gold.
    \item Partial matches for multi-part answers.
\end{itemize}

\textbf{\boxed{C} - INCOMPLETE/REPETITIVE/REFUSAL:}
\begin{itemize}[noitemsep, topsep=0pt]
    \item Invalid answers (must specify error type).
\end{itemize}

\textbf{Annotation Criteria:}
\begin{enumerate}[noitemsep, topsep=0pt]
    \item First check validity (INCOMPLETE, REPETITIVE, REFUSAL). 
    \item Compare candidate’s final answer vs. gold answer in detail. 
        \begin{itemize}[noitemsep, topsep=0pt]
            \item Identify strict requirements (e.g., exact match, order, completeness).
            \item Allow tolerances (format differences, equivalent math forms, unsimplified fractions, provide the full answer for completion-type questions). 
            \item Check for equivalences, e.g., 
            $\frac{2x-7}{(x+1)(x-2)} \text{ and } \frac{3}{x+1} - \frac{1}{x-2}$\ are equivalent.
            
            \item Consider:
            \begin{itemize}
                \item Factoring/expansion: $x^2+2x+1 \to (x+1)^2$
                \item Fraction simplification: $\frac{x^2-1}{x+1} \to x-1$
                \item Partial fraction decomposition, decimals, trig identities, substitutions, etc.
            \end{itemize}
            \item For multiple-choice, answer must exactly match gold or be fully equivalent.
        \end{itemize}
    \item Only give the final judgment.
\end{enumerate}

\textbf{Final Judgment:} A/B/C

Here is your task:

\textless Original Question Begin\textgreater
\{question\}
\textless Original Question End\textgreater

\textless Standard Answer Begin\textgreater
\{gold answer\}
\textless Standard Answer End\textgreater

\textless Candidate's Answer Begin\textgreater
\{llm response\}
\textless Candidate's Answer End\textgreater

Final Judgment:

\end{namedbox}
\end{center}
\begin{center}
\begin{namedbox}[width=0.9\linewidth, fontupper=\small,label=other_inference]{Prompt of Inference for the evaluation of robustness}
You are a diligent and precise assistant tasked with evaluating the correctness of responses. You will receive a question, an output sentence, and the correct answer. Your task is to determine if the output sentence accurately answers the question based on the provided correct answer. Respond with either [Correct] or [Incorrect].\\
-
Special considerations:\\

1. **Multiple Answers**: If the output contains multiple answers, evaluate whether later answers modify or correct earlier ones. In such cases, compare the final answer with the correct answer. If the final answer is unclear or incorrect, respond with [Incorrect].\\

2. **Mathematical Problems**: If the formats differ but the answers are mathematically equivalent, respond with [Correct].\\

3. **Explicit Options**: If the question provides explicit candidate answers, the output will be considered correct if it clearly indicates the correct option's code or the correct option's content.\\

4. **No Explicit Options**: If the question does not provide explicit options, the output must align with the correct answer in content and meaning to be considered [Correct].\\
-

Question: \{question\}

Output sentence: \{llm response\}

Correct answer: \{gold answer\}

Judgement:

\end{namedbox}
\end{center}
\subsection{Data Details} \label{sec:appendix:statics:details}
In this section, we present several data cases, focusing primarily on examples of the equivalent forms we generated. Box.~\ref{box:math_sample} shows an equivalent example in mathematics, where the answers in the Outputs will undergo both LLM annotation and human annotation. Box.~\ref{box:phy_sample} presents an equivalent example in physics, Box.~\ref{box:chem_sample} in chemistry, Box.~\ref{box:bio_sample} in biology, and Box.~\ref{box:qa_sample} provides an example for QA questions.

\begin{center}
\begin{namedbox}[label=box:math_sample]{A sample for generating equivalent answers to mathematical problems}
\textbf{Question}:
Find all functions 
$ f: \mathbb{Q}^{+} \mapsto \mathbb{Q}^{+}$ 
such that: 
\begin{center}
$ f(x) + f(y) + 2xy f(xy) = \frac{f(xy)}{f(x+y)}.$
\end{center}


\textbf{Gold Ans}: 
$\frac{1}{x^2}$

\textbf{Outputs:}

\hspace*{4mm}
$\boxed{{x^{-2}}}$

\hspace*{4mm}
$\boxed{{\frac{1}{x \cdot x}}}$

\hspace*{4mm}
$\boxed{{\left(\frac{1}{x}\right)^2}}$

\hspace*{4mm}
$\boxed{{\frac{1}{x^2} \text{ for } x \ne 0}}$

\hspace*{4mm}
$\boxed{{\frac{1}{x^2} = x^{-2}}}$

\end{namedbox}
\end{center}

\begin{center}
\begin{namedbox}[label=box:phy_sample]{A sample for generating equivalent answers to physical problems}
\textbf{Question}:
Set a crystal prism with an apex angle of $60^{\circ}$, and the optical axis is perpendicular to the main cross-section of the prism. The sodium yellow light is refracted in the prism in the direction of the minimum deviation angle and focused using a lens with a focal length of 1 m. What is the separation between the o-ray and e-ray spectral lines?

\textbf{Gold Ans:} $7.16 \text{ mm}$

\textbf{Outputs:}

\hspace*{4mm}
$\boxed{{7.16 \, \text{mm}}}$

\hspace*{4mm}
$\boxed{{7.16 \, \text{mm}}}$

\hspace*{4mm}
$\boxed{{0.00716 \, \text{m}}}$

\hspace*{4mm}
$\boxed{{7160 \, \mu\text{m}}}$

\hspace*{4mm}
$\boxed{{7.16 \times 10^{-3} \, \text{m}}}$

\end{namedbox}
\end{center}

\begin{center}
\begin{namedbox}[label=box:chem_sample]{A sample for generating equivalent answers to chemical problems}
\textbf{Question}:
I'm looking for the SMILES version of this IUPAC chemical name. 

\begin{center}

N4-(3,4-difluorophenyl)-2-methyl-N6-prop-2-enylpyrimidine-4,6-diamine
\end{center}

\textbf{Gold Ans}: CC1=NC(=CC(=N1)NC2=CC(=C(C=C2)F)F)NCC=C

\textbf{Outputs:}

\hspace*{4mm}
$\boxed{\text{N4-(3,4-difluorophenyl)-2-methyl-N6-prop-2-enylpyrimidine-4,6-diamine}}$

\hspace*{4mm}
$\boxed{\text{2-methyl-N6-(prop-2-en-1-yl)-N4-(3,4-difluorophenyl)pyrimidine-4,6-diamine}}$

\hspace*{4mm}
$\boxed{\text{CC1=NC(=CC(=N1)NC2=CC(=C(C=C2)F)F)NCC=C}}$

\hspace*{4mm}
$\boxed{\text{C12N(C=C(N1)C)C=CC(NC3=CC=C(C=C3)F)C=C2F}}$

\end{namedbox}
\end{center}
\clearpage

\begin{center}
\begin{namedbox}[label=box:bio_sample]{A sample for generating equivalent answers to biological problems}
\textbf{Question:}
Please predict the dot-bracket notation of the secondary structure directly from the RNA sequence: 

\begin{center}
{\tiny{UCUGAUCCAUGAUGACUUCCAAACACAUGUGGACUGUUGGAUGAAGCAUGCGGAUAUGGGACUGAGAUCAGAC}}
\end{center}

\textbf{Gold Ans:} 
.((((..............................................................))))..

\textbf{Outputs:}

\hspace*{4mm}
\begin{center}
$\boxed{\text{.((((..............................................................))))..}}$
\end{center}

\hspace*{4mm}
\begin{center}
\boxed{\text{CT: 2 30, 3 29, 4 28, 5 27, 6 26, 7 25, 8 24, 9 23, 10 22, 11 21, 12 20, 13 19, 14 18, 15 17}}
\end{center}

\begin{equation*}
\boxed{
\begin{aligned}
& \text{BPseq:}\\ 
& \text{2 30, 3 29, 4 28, 5 27, 6 26, 7 25, 8 24, 9 23, 10 22, 11 21, 12 20, 13 19, 14 18, 15 17}
\end{aligned}
}
\end{equation*}

\begin{equation*}
\boxed{
\begin{aligned}
&\text{pairs=}\\ 
&\text{[(2,30),(3,29),(4,28),(5,27),(6,26),(7,25),(8,24),}\\
&\text{(9,23),(10,22),(11,21),(12,20),(13,19),(14,18),(15,17)]}
\end{aligned}
}
\end{equation*}

\begin{equation*}
\boxed{
\begin{aligned}
&\text{matrix:} \\ &\text{[[0,0,0,0,0,0,0,0,0,0,0,0,0,0,0,0,0,0,0,0,0,0,0,0,0,0,0,0,0,0,0],} \\
&\text{[0,0,0,0,0,0,0,0,0,0,0,0,0,0,0,0,0,0,0,0,0,0,0,0,0,0,0,0,0,0,1],}\\
&\text{..., }\\
&\text{[0,1,0,0,0,0,0,0,0,0,0,0,0,0,0,0,0,0,0,0,0,0,0,0,0,0,0,0,0,0,0]]}
\end{aligned}
}
\end{equation*}
\end{namedbox}
\end{center}

\begin{center}
\begin{namedbox}[label=box:qa_sample]{A sample for generating equivalent answers to QA problems}
\textbf{Question}:
What's the name of the active president of the United States in Season 1 of Motherland: Fort Salem?

\textbf{Gold Ans}: President Kelly Wade

\textbf{Outputs:}

\hspace*{4mm}
\begin{equation*}
\boxed{
\begin{aligned}
&\text{The active president of the United States in Season 1}\\
&\text{of Motherland: Fort Salem is named Kelly Wade.}
\end{aligned}
}
\end{equation*}

\hspace*{4mm}
$\boxed{\text{Kelly Wade is U.S. President in Season 1 of Motherland: Fort Salem.}}$

\hspace*{4mm}
$\boxed{\text{In the first season of Motherland: Fort Salem, the president is Kelly Wade.}}$

\hspace*{4mm}
$\boxed{\text{Kelly Wade is the incumbent U.S. leader in Season 1 of Motherland: Fort Salem.}}$

\hspace*{4mm}
$\boxed{\text{In Season 1 of Motherland: Fort Salem, Kelly Wade serves as the U.S. president.}}$

\end{namedbox}
\end{center}

\section{Details of Training}\label{sec:appendix:train}
In this section, we present the parameter configurations used during the training process. The training mainly involves two parts: SFT and RL. For SFT, we adopt full fine-tuning, and the detailed parameter configurations are shown in Tab.~\ref{tab:appendix:sft}. For RL, a modified version of GRPO is used, with detailed parameters also provided in Tab.~\ref{tab:appendix:rl}.

\section{Limitations and Future Work}\label{sec:appendix:future}
In this section, we discuss the limitations of our work and outline directions for future research. We propose a verifier for scientific verification tasks that demonstrates strong reasoning capabilities, achieving high performance with concise and interpretable reasoning outputs. However, some scenarios demand both high accuracy and extreme efficiency. In future work, we plan to leverage the model’s explicit reasoning abilities to further enhance its implicit reasoning, allowing it to maintain strong performance even without explicitly generating detailed reasoning steps. This approach could provide significant efficiency gains while preserving the model’s reliability and robustness across a wider range of scientific verification tasks.
\begin{table}[htbp]
\centering
\caption{SFT Configuations for SCI-Verifier.}
\label{tab:appendix:sft}
\vspace{-8pt}
\begin{tabular}{l l}
\toprule
\textbf{Parameter} & \textbf{Value} \\
\midrule
\midrule
BF16 & True \\
Gradient Checkpointing & False \\
Learning Rate & \num{5e-5} \\
LR Scheduler Type & cosine\_with\_min\_lr \\
Minimum LR Rate & 0.1 \\
Packing & False \\
Maximum Sequence Length & 1024 \\
Maximum Steps & -1 \\
Number of Training Epochs & 2 \\
Per Device Train Batch Size & 2 \\
Per Device Eval Batch Size & 16 \\
GPUs Per Node & 4 \\
Number of Nodes & 1 \\
Seed & 42 \\
Use Liger Kernel & True \\
Warmup Ratio & 0.02 \\
\bottomrule
\end{tabular}
\vspace{-12pt}
\end{table}
\begin{table}[htbp]
\centering
\caption{RL Configuations for SCI-Verifier.}
\vspace{-8pt}
\label{tab:appendix:rl}
\begin{tabular}{l l}
\toprule
\textbf{Parameter} & \textbf{Value} \\
\midrule
\midrule
BF16 & True \\
Temperature & 1.0 \\
Top p & 1.0 \\
Clip Ratio Low & 0.2 \\
Clip Ratio High & 0.28 \\
Max Response Length & 2048 \\
Overlong Buffer Len & 1024 \\
Learning Rate & \num{1e-6} \\
Number of Training Epochs & 20 \\
GPUs Per Node & 4 \\
Number of Nodes & 1 \\
Seed & 42 \\
\bottomrule
\end{tabular}
\vspace{-12pt}
\end{table}

\section{The Use of Large Language Models}\label{sec:appendix:usage}
In our paper, LLMs are first used to polish the writing, improving the clarity and readability of the manuscript. At the same time, as mentioned multiple times in the main text and Appendix, we employ LLMs to generate and annotate training and test data. Since LLM outputs can sometimes be unreliable, as noted in the text, all selected data are subsequently manually re-annotated by human experts and carefully filtered.

\end{document}